%% file: paper.tex
\documentclass[times,twocolumn,final]{elsarticle}

\usepackage{framed,multirow}

\usepackage{latexsym}

\usepackage{graphicx}
\usepackage{amsmath}
\usepackage{amssymb}
\usepackage{color}
\usepackage[table]{xcolor}
\usepackage[lofdepth,lotdepth, caption=false,font=footnotesize]{subfig}
\usepackage{hhline}
\usepackage{stfloats}
\definecolor{Gray}{gray}{0.9}
\usepackage[utf8]{inputenc}
\usepackage{hyperref}
\usepackage{soul}
\usepackage{array}
\usepackage{booktabs}
\usepackage{multicol}
\usepackage{multirow}
\usepackage{tabularx}
\usepackage{threeparttable}
\usepackage{bigdelim}
\usepackage{float}
\usepackage{caption}
\usepackage{makecell}
\usepackage{diagbox}
\usepackage{physics}
\usepackage{tikz-cd}
\usepackage{adjustbox}

\input{figs}

\input{tabs}

\newcommand{\figref}[1]{Fig.~\ref{fig:#1}}
\newcommand{\tblref}[1]{Table~\ref{tbl:#1}}
\newcommand{\secref}[1]{Section~\ref{sec:#1}}

\newcommand{\ra}[1]{\renewcommand{\arraystretch}{#1}} 

\begin{document}


\begin{frontmatter}

\title{Subtle Signals: Video-based Detection of Infant Non-nutritive Sucking \\as a Neurodevelopmental Cue}


\author[1]{Shaotong Zhu}
\author[1]{Michael Wan}
\author[1]{Sai Kumar Reddy Manne}
\author[2]{Emily Zimmerman}
\author[1]{Sarah Ostadabbas \corref{cor1}}
\cortext[cor1]{Corresponding author:   Email: ostadabbas@ece.neu.edu}

\address[1]{Augmented Cognition Lab, Department of Electrical and Computer Engineering, Northeastern University, Boston,
MA, USA}

\address[2]{Speech \& Neurodevelopment Lab, Department of Communication Sciences and Disorders, Northeastern University, Boston,
MA, USA}


\begin{abstract}
Non-nutritive sucking (NNS), which refers to the act of sucking on a pacifier, finger, or similar object without nutrient intake, plays a crucial role in assessing healthy early development. In the case of preterm infants, NNS behavior is a key component in determining their readiness for feeding. In older infants, the characteristics of NNS behavior offer valuable insights into neural and motor development. Additionally, NNS activity has been proposed as a potential safeguard against sudden infant death syndrome (SIDS). However, the clinical application of NNS assessment is currently hindered by labor-intensive and subjective finger-in-mouth evaluations. Consequently, researchers often resort to expensive pressure transducers for objective NNS signal measurement. To enhance the accessibility and reliability of NNS signal monitoring for both clinicians and researchers, we introduce a vision-based algorithm designed for non-contact detection of NNS activity using baby monitor footage in natural settings. Our approach involves a comprehensive exploration of optical flow and temporal convolutional networks, enabling the detection and amplification of subtle infant-sucking signals. We successfully classify short video clips of uniform length into NNS and non-NNS periods. Furthermore, we investigate manual and learning-based techniques to piece together local classification results, facilitating the segmentation of longer mixed-activity videos into NNS and non-NNS segments of varying duration. Our research introduces two novel datasets of annotated infant videos, including one sourced from our clinical study featuring 19 infant subjects and 183 hours of overnight baby monitor footage. Additionally, we incorporate a second, shorter dataset obtained from publicly available YouTube videos. Our NNS action recognition algorithm achieves an impressive 95.8\% accuracy in binary classification, based on 960 2.5-second balanced NNS versus non-NNS clips from our clinical dataset. We also present results for a subset of clips featuring challenging video conditions. Moreover, our NNS action segmentation algorithm achieves an average precision of 93.5\% and an average recall of 92.9\% across 30 heterogeneous 60-second clips from our clinical dataset. 
\end{abstract}


\end{frontmatter}

\section{Introduction}
\label{sec:intro}

Infant feeding requires a delicate harmony between sucking, swallowing, and breathing movements, often presenting a challenge for newborn and especially preterm infants: around 2.8 million infants in all face feeding challenges per year in the U.S. Nurse clinicians often gauge feeding readiness with subjective finger-in-mouth assessments of \textit{non-nutritive sucking} (NNS)---sucking without nutrient delivery---but this can cause discomfort or lead to serious complications if the assessment is mistaken \cite{benjasuwantep_feeding_2013}. An automated, objective, video-based tool for tracking infant NNS would help address these concerns, and pave the way for a fully automated contactless feeding assessment system in the future. Aside from aiding clinical decision-making in real-time, such a tool could also benefit research in infant neurodevelopmental diagnostics. Given the limited range of motor function and means of expression in infancy, characteristics of NNS constitute critical signals of neural and motor development in early life \cite{medoff-cooper_neonatal_1995}, and NNS has even been proposed as a potential mechanism for reducing the risk of sudden infant death syndrome (SIDS) \cite{psaila_infant_2017,zavala_abed_how_2020}, the leading cause of death of US infants aged between 1 and 12 months \cite{carlin2017risk}. Understanding the relation between NNS patterns and characteristics of breathing, feeding, and arousal during sleep could enhance scientific understanding of infant neurodevelopment and protective factors for SIDS. Nonetheless, few such studies have been conducted, partly due to the difficulty of measuring the NNS signal.

Non-nutritive sucking typically manifests in bursts comprising approximately 6 to 12 sucks, occurring at a rate of 2 Hz per suck. These bursts sporadically appear a few times per minute during periods of heightened non-nutritive sucking activity, as noted in Zimmerman et al.'s study \cite{zimmerman_changes_2020}. Nevertheless, active non-nutritive sucking phases can be infrequent, often constituting only a few minutes per hour. This intermittent nature of non-nutritive sucking imposes a substantial workload on clinicians and researchers seeking to investigate its characteristics and how it evolves over time. Current transducer-based methodologies, as illustrated in \figref{figPacDevice}, effectively monitor non-nutritive sucking activity \cite{zimmerman_patterned_2017}. However, these approaches are associated with high costs, limited suitability for research purposes, and potential interference with the natural sucking behavior itself. This compelling context drives our initiative to develop an end-to-end computer vision system tailored for the recognition and segmentation of infant non-nutritive sucking actions in videos recorded overnight within natural settings. Our primary objective is to facilitate broad applications in automatic screening and telehealth. We place a strong emphasis on achieving high precision, ensuring the reliable extraction of periods of sucking activity for subsequent analysis by human experts.

\figPacDevice

Our technical contributions encompass two key aspects: firstly, addressing the fine-grained NNS action recognition challenge, which involves classifying 2.5-second video clips into NNS or non-NNS categories; and secondly, tackling the broader NNS action segmentation problem, which entails identifying frames that exhibit NNS activity in minute-long video clips. Our action recognition method relies on spatiotemporal learning through convolutional long short-term memory networks. To overcome the limitations posed by the scarcity and reliability issues of real-world baby monitor footage, our pipeline incorporates a specialized infant pose state estimation technique. This method detects the infant's face, narrows the focus to the mouth and pacifier region, and enhances it using dense optical flow. For action segmentation, we explore both manually-tuned and learning-based approaches for aggregating and filtering the outcomes of local NNS recognition. Our methodology serves as the foundation for a fully automated computer vision assessment of NNS, enabling the extraction of critical sucking signal characteristics, including frequency, duration, amplitude, and temporal pattern.

\figGallery
We present two new datasets in our work: the \textbf{NNS clinical in-crib dataset}, consisting of 183 hours of nighttime in-crib baby monitor footage collected from 19 infants and annotated for NNS activity and pacifier use by our interdisciplinary team of behavioral psychology and machine learning researchers, and the \textbf{NNS in-the-wild dataset}, consisting of 10 naturalistic infant video clips annotated for NNS activity. \figref{nnsSignal} displays sample frames from both datasets.

This article is an extension of previous work of ours \cite{zhu2023video}. Our main contributions, with \textit{new work} emphasized, include\footnote{Our code and the manually annotated NNS in-the-wild dataset can be found at \url{https://github.com/ostadabbas/NNS-Detection-and-Segmentation}.}: 

\begin{itemize}
    \item the creation of the first infant video datasets manually annotated with NNS activity, \textit{including an additional subset with clips featuring challenging infant poses, motions, and conditions};
    \item the development of an NNS classification system using a convolutional long short-term memory network, aided by infant domain-specific face localization, video stabilization, and customized signal enhancement, \textit{with new performance tests on the challenging dataset};
    \item \textit{an exhaustive experimental comparison of our classification method with various spatiotemporal models}; and
    \item successful NNS segmentation on longer clips by aggregating local NNS recognition, both with a manually-tuned sliding windows approach, \textit{and a deep-learning based approach using a dilated convolutional network}.
\end{itemize}

\section{Related Work}
Our work is a novel approach to non-nutritive sucking (NNS) detection using tools from computer vision human action and recognition. We review prior, largely contact-based approaches to NNS detection in \secref{related-nns}, and also general computer vision methods for human action recognition and segmentation in \secref{related-action-rec} and \secref{related-action-seg} relevant to our purposes.


\subsection{Non-Nutritive Sucking Detection}

\label{sec:related-nns}

The primary method for acquiring non-nutritive sucking data involves using a pressure-sensor-equipped pacifier. This sensor, as described in \cite{zimmerman_patterned_2017}, is pivotal in detecting and quantifying infant NNS patterns during pacifier use. Typically, the sensor is housed within the pacifier handle or a separate unit. INNARA HEALTH developed the NTrainer System \cite{poore2008patterned} to enhance NNS and feeding development in premature and newborn infants. They employ the Actifier, a specialized system using a Honeywell pressure transducer integrated with a custom Delrin receiver and a sterile smoothie silicone pacifier, to measure lip, tongue, and jaw forces during sucking. However, these traditional methods are prohibitively expensive, and also potentially alter the infant sucking patterns they are trying to measure.

\cite{huang2019infant}, from the same lab as the corresponding author, presents a novel contactless method for collecting NNS data.
This approach automatically tracks the baby's jaw landmarks in video footage via 2D facial landmarks, then employs a 3D morphable model (3DMM) \cite{huber2016multiresolution} to generate 3D facial landmarks. 
Then suck cycles and NNS pattern frequency are computed from the denoised landmark movement signals. However, the 3DMM model is only learned from adult face data, limiting its accuracy, given the domain gap between infant and adult faces (see \cite{wan_infanface_2022}). The overall pipeline is inference-based, with no component trained on infant or NNS data, and is only tested on 10 short video samples without NNS annotations. Our current work significantly expands on our study of NNS detection, specifically targeting the tasks of NNS action recognition and segmentation. by training and testing a learned pipeline on datasets which in aggregate included over 1000 clips from dozens of infant subjects.

\subsection{Human Action Recognition}
\label{sec:related-action-rec}
Action recognition or action classification, used interchangeably for our purposes, is the task of assigning a class label from a fixed list to a short video clip. The actions are typically short and well-defined, like riding a bike or climbing stairs. Datapoints consist of short video clips, often on the order of a few seconds long, trimmed to contain an single unequivocal action. Leveraging the success of 2D convolutional neural networks (CNNs) in image analysis, many action recognition methods have been built upon this robust CNN architecture.
Many existing video-based action recognition models are simply built on top of image classification models, which are tailored to process video by replacing 2D- with 3D convolution, such as 3D ResNet \cite{kataoka2020would} that extended the success of 2D convolutional networks to three-dimensional spatiotemporal data, laying the foundation for video understanding. To better address the temporal queue yet preserve the spatial feature, I3D \cite{carreira2017quo} introduced a pivotal concept by fusing information from two streams: RGB and optical flow, using two 2D models with identical network structures, thereby enhancing action recognition performance through an integrated approach. Furthermore, X3D \cite{feichtenhofer2020x3d}, has made significant progress towards efficient video architecture that presents new insights for turning a 2D architecture into a 3D one by progressively expanding it along multiple axes, such as width, depth, and time. 

A limitation is evident in the aforementioned CNN-based methodologies pertains to their predominant application in addressing coarse-grained action recognition tasks (e.g., playing golf or tennis etc.), wherein they have demonstrated remarkable performance primarily attributable to the pre-training of their 2D base models on large-scale coarse-grained image datasets like ImageNet  (\cite{deng2009imagenet}). In our particular research context, our objective is the classification of short video clips depicting infants based on the subtle presence or absence of non-nutritive sucking (NNS) behavior---a nuanced facial action characterized by minute movements around the mouth region. Despite our endeavors to adapt the aforementioned approaches for NNS action classification (as reported in \secref{segmentation-results}), we encountered a pronounced performance degradation in comparison to their proficiency in coarse-grained action recognition, primarily due to the presence of a substantial action domain gap. In response to the subtlety inherent to NNS actions, we adopted an approach akin to I3D, leveraging optical flow input to account for minuscule motion patterns. Subsequently, we expanded upon this framework by incorporating 2D convolutional neural networks into the temporal dimension, allowing for the processing of spatiotemporal data. This augmentation involved the integration of sequential networks, specifically long short-term memory (LSTM) networks, subsequent to frame-wise convolutions, thereby fortifying the model's ability to capture medium-range temporal dependencies, as elucidated by \cite{yue2015beyond}.

\subsection{Human Action Segmentation}
\label{sec:related-action-seg}

Temporal action segmentation is a broader task in video comprehension. The goal is to take a longer video consisting of a diverse spectrum of activities, partition it into a set of intervals of arbitrary duration in time, and assign action classes to each interval. Recent advancements in this domain have predominantly adopted the multiple-instance learning (MIL) paradigm \cite{maron1997framework}, wherein the entirety of an untrimmed video is conceptualized as a labeled bag encompassing numerous unlabeled instances. Within this framework, a common approach involves the treatment of video snippets as individual instances, utilizing a pre-trained feature extractor rooted in action recognition models. This feature extractor is employed in conjunction with a sliding window mechanism to construct an input feature sequence, which is subsequently used to train a segmentation model tasked with classifying the labels associated with the snippets within the sequence, ultimately enabling the precise segmentation of actions within the video.

Following the MIL paradigm, the MS-TCN \cite{farha2019ms} pioneered the concept of multi-stage temporal convolutions, offering a hierarchical framework for capturing long-range temporal dependencies by processing video sequences in a multi-scale fashion. Building upon this foundation, Global2Local \cite{gao2021global2local} introduced an innovative perspective by integrating global and local context modeling, enhancing the network's ability to discern intricate spatiotemporal patterns. The logical continuum culminates with ASFormer \cite{yi2021asformer}, where the transformer architecture is adapted to spatiotemporal video data so that the strengths of transformers in capturing global context are leveraged but maintaining local spatiotemporal information through tokenization strategies, thereby bridging the gap between global and local representations. 

The aforementioned methodologies all incorporated a pre-trained I3D-based feature extractor as a preliminary step for feature sequence preparation in their training pipelines. In our pursuit of investigating the generalization potential of our newly proposed action recognition model, in contrast to the previously suggested aggregation-based model \cite{zhu2023video}, we undertook a modification of the MS-TCN model. Subsequently, we conducted an extensive evaluation employing the features extracted by our pre-trained action recognition model. To discern the efficacy and comparative performance of this model against the other state-of-the-art methods, all of which were fine-tuned on the identical set of features, a comparative analysis was conducted as elaborated in \secref{other-segmentation-methods}. This rigorous evaluation aims to shed light on the suitability and performance characteristics of our deep learning-based model in contrast to previously advocated aggregation-based approaches, contributing to a more comprehensive understanding of action recognition in the context of video analysis.

\section{NNS datasets}
\label{sec:dataset}

\setup
\annotSoft
\subsection{Data Collection and Annotation}
\label{sec:data-preparation}

Our primary dataset is the \textbf{NNS clinical in-crib dataset}, collected using the toolkit shown in \figref{setup} consisting of 183 hours of baby monitor footage collected from 19 infants during overnight sleep sessions by our clinical neurodevelopment team, with Institutional Review Board (IRB \#17-08-19) approval. Videos were shot in-crib with the baby monitors set up by caregivers, under low light triggering the monochromatic infrared mode. Tens of thousands of timestamps for NNS and pacifier activity were placed using the annotation tool shown in \figref{annotSoft}, by two trained behavioral coders per video. For NNS, the definition of an event segment was taken to be an NNS \textit{burst}: a sequence of sucks with ${<}1$ s gaps between. We restricted our subsequent study to NNS during pacifier use, which was annotated more consistently. Cohen $\kappa$ annotator agreement of NNS events during pacifier use (among 10 pacifier-using infants) averaged 0.83 in 10 s incidence windows, indicating strong agreement by behavioral coding standards, but we performed further manual selection to increase precision for machine learning use,
We also created a smaller but publicly available \textbf{NNS in-the-wild dataset} of 14 YouTube videos featuring infants in natural conditions, with lengths ranging from 1 to 30 minutes, and similar annotations.

\subsection{NNS Clinical In-Crib Dataset Statistics}

\tblcohenkappa

To demonstrate the reliability of our annotations, we provide Cohen $\kappa$ inter-rater reliability scores for each pair of annotators' behavioral coding, per infant video. Since annotators may not agree on the number of events in any given period, Cohen $\kappa$s cannot be computed directly on the timestamp data. Instead, we adhere to common practice from behavioral coding in psychology and convert each coder's annotations for a single event type (NNS or pacifier) in a video to a binary time sequence representing uniform windows in the runtime, with with 1 assigned to windows which overlap temporally with at least one event of that type, and 0 assigned to the remaining windows. We consider both the fine-grained windows of 0.1 s, which contain one video frame each, as well as the coarser windows of 10 s, which is more in line with conventions in behavioral coding, given the imprecision and differences in interpretation built into human behavioral assessments. The Cohen $\kappa$ scores\footnote{The Cohen $\kappa$ agreement between two raters' binary classifications on a set is defined as $\kappa:=\frac{p_\text{o}-p_\text{e}}{1-p_\text{e}}$, where $p_\text{0}$ is the observed portion of agreements in the set and $p_\text{e}$ is the estimated probability of chance agreement, itself defined by $p_\text{e}:=p_0 p_1 + (1-p_0)(1-p_1)$, with $p_0$ and $p_1$ being the positive assignment rate for each respective rater. It is intended to measure the level of agreement between two raters' assessments while taking into account chance agreements. We adopt the following suggested interpretations of agreement strength based on $\kappa$ score from \cite{mchugh_interrater_2012}: 0--0.2 means no agreement, 0.21--0.39 minimal, 0.40--0.59 weak, 0.60--0.79 moderate, 0.80--0.90 strong, and $>$0.90 almost perfect.} for the events considered as time sequence over both 0.1 s and 10 s windows, aggregated across all infants in our training and test data, is reported in \tblref{cohen-kappa}. In addition to raw NNS and pacifier events, the table also shows agreement for the derived annotation of NNS events occurring only during pacifier events. Such NNS action is far more regular and reliably codable, and hence we restrict our video segmentation efforts to those events alone. The interpretation of $\kappa$ scores is subjective, but the levels achieved by the pacifier annotations would typically be characterized as indicating ``near perfect'' agreement; the NNS-with-pacifier annotation scores could be considered ``weak'' or ``moderate'' agreement under the harsh 0.1 s intervals, and ``strong'' or ``almost perfect'' under the 10 s intervals. Given the inherent difficulty of NNS annotation, the sheer amount of runtime of the video data, and our subsequent success in using the data for the segmentation task, we believe these annotation efforts represent a hard-earned success. 

\tblnnsstats

\tblref{nns-stats} displays statistics derived from our NNS and pacifier annotations for our 10 subjects, with the NNS events restricted to those annotated during pacifier events (according to the same annotator), based on scientific interests. As expected, there is wide variation in both NNS and pacifier event count and average duration per subject, with for instance the NNS count ranging from 2.3 to 60.4 per hour, and NNS duration from 0.1 to 4.5 minutes per hour. The average length of an NNS event (a burst of sucks) per subject is somewhat more uniform, ranging from 3.2 to 8.1 seconds. 

\subsection{Dataset Clip Curation}

From our hours-long annotated footage, we curate the following reference datasets to support classification and segmentation tasks, guided by the above reliability and statistical considerations. While our NNS annotations can be considered strongly reliable based on behavioral coding standards, further filtering is necessary to reach sufficient reliability on the split-second level typically desirable in machine learning. But given rarity of NNS activity (0.1--4.5 min/h), positive examples have to be over-represented in order to provide sufficient data for training or support statistically significant conclusions for testing. 

From each of our NNS in-crib and in-the-wild datasets, we extracted 2.5 s clips for the classification task and 60 s clips for the segmentation task. In the NNS clinical in-crib dataset, we restricted our attention to six infant videos containing enough NNS activity during pacifier use for meaningful clip extraction. From each of these, we randomly drew up to 80 2.5 s clips consisting entirely of NNS activity and 80 2.5 s clips containing non-NNS activity for classification, for a total of 1,600; and five 60 s clips featuring transitions between NNS and non-NNS activity for segmentation, for a total of 30; redrawing if available when annotations were not sufficiently accurate. In the NNS in-the-wild dataset, we restricted to five infants exhibiting sufficient NNS activity during pacifier use, from which we drew 38 2.5 s clips each of NNS and no NNS activity for classification, for a total of 76; and from two to 26 60 s clips of mixed activity from each infant for segmentation, for a total of 39; again redrawing in cases of poor annotations. 

During the annotation process of our NNS in-crib dataset, we encountered several cases of NNS activity that were hard to distinguish from non-NNS activity, primarily due to the background movements, such as the infant's crib swinging in the video frame. To enable a specific study of such tricky scenarios, we isolated a new \textbf{challenging subset} of our NNS clinical in-crib dataset, consisting of 120 2.5 s videos drawn evenly from our six final subjects. Training and testing on this dataset, as we do in \secref{performance-challenging}, gives a broader sense of performance under difficult real-world conditions.


\section{Method}
\label{sec:methods}
\figDiagram

Our two-stage NNS action segmentation pipeline, shown in \figref{Diagram}, is designed to process extended videos featuring infants using pacifiers and predict the timestamps at which NNS events occur throughout the entire video. Input videos of arbitrary length are organized into shorter segments via sliding windows, 2.5 s in length. In the first stage, the 2.5 s windows are classified into NNS or non-NNS classes using our NNS action recognition module, described in \secref{action-recognition}. In the second stage, these classification signals are amalgamated to generate a segmentation outcome for the whole video, consisting of a list of start and end timestamps for NNS events. This action segmentation module is described in \secref{action-segmentation}. Here, we focus on general methods, and leave specific implementation details such as neural network model types to \secref{experiments}.

\subsection{NNS Action Recognition}
\label{sec:action-recognition}

Our action recognition module includes a frame-based preprocessing step, followed by analysis via a spatiotemporal neural network. The preprocessing includes the following transformations in sequence. All three steps are used to produce training data for the subsequent spatiotemporal classifier, but during inference, the data augmentation step is not applicable and is omitted.

\begin{description}
    \item[Smooth facial crop] We use the RetinaFace face detector \cite{retinaface2019} to analyze frames within each video clip until a face bounding box is located. This bounding box is then propagated to adjacent frames using the Minimum Output Sum of Squared Error (MOSSE) tracker \cite{mossepaper}. To enhance the consistency of the facial bounding box sequence and mitigate temporal gaps, we identify saliency corners \cite{shi1994good} in the initial frame and track them to the subsequent frame employing the Lucas-Kanade optical flow algorithm \cite{lucas1981iterative}. We further enhance the trajectory's smoothness by applying a moving average filter and then apply this trajectory to each bounding box, thereby stabilizing the facial region. Finally, we crop the raw input video using this smoothed bounding box, resulting in a video featuring only the face.
    \item[Data augmentation] During the video preprocessing stage, as part of training data generation for the spatiotemporal classifier, we introduce random transformations to the face-cropped video. These transformations include actions like rotations, scaling adjustments, and flipping. This augmentation process aims to enhance the model's generalizability, especially in scenarios where we have limited data available.
    \item[Optical flow] \opFlow Following the trimming and augmentation steps, we compute the short-time dense optical flow \cite{liu2009beyond} between consecutive frames. We then transform the optical flow results into the Hue, Saturation, and Value (HSV) color space by combining the optical flow direction vector and the magnitude of each pixel. This process accentuates the visible motion between frames, amplifying subtle NNS movements, as demonstrated in \figref{opFlow}.\footnote{Through informal qualitative evaluations, we ascertained that dense optical flow outperforms alternative implementations like Farneback \cite{farneback2003two}, TV-L1 \cite{pock2007duality}, and RAFT \cite{teed2020raft}.}
\end{description}

After these preprocessing steps, the resulting optical flow video frames are passed to a spatiotemporal module, which predicts an action class label (either NNS or non-NNS). The structure of our spatiotemporal module is a 2D--1D convolutional network: individual frames are passed into a conventional (2D) convolutional neural network, and the resulting spatial features for each frame are passed into a temporal (1D) convolution network for final classification. In our experiments (see \tblref{compareClassification}), this worked more effectively than two-stream or 3D convolutional methods. See \secref{nnsResults} for more on specific network choices.


\subsection{NNS Action Segmentation}
\label{sec:action-segmentation}

We explore two types of methods for amalgamating local NNS action recognition outcomes into a global NNS action segmentation result, the first based on simple aggregations of the local classification results, and the second a learned model which uses the features generated by the local classifier.

Our aggregation methods work directly with the binary classification results on the 2.5 s sliding windows. This window size---26 frames of the 10 Hz footage---was chosen to be small enough to allow for relatively fine-grained segmentation results, while at the same time large enough to allow some flexibility for human annotation subjectivity and variation in reaction time. By working with sliding windows with 0.5 s strides, we can still produce segmentation results with 0.5 s effective resolution. These considerations lead naturally to the following three aggregation methods:

\begin{description}
    \item[Tiled] 2.5 s windows precisely tile the length of the video with no overlaps, and the classification outcome for each window is taken directly to be the segmentation outcome for that window.
    \item[Sliding] 2.5 s windows are slid across with 0.5 s strides, and the classification outcome for each window is assigned to its (unique) middle-fifth 0.5 s segment as the segmentation outcome.
    \item[Smoothed] 2.5 s windows are slid across with 0.5 s strides, the classification \textit{confidence score} for each window is assigned to its middle-fifth 0.5 s segment, a 2.5 s moving average of these confidence scores are taken, then the averaged confidence scores are thresholded for the final segmentation outcome.
\end{description}

We turn to our learned action segmentation model. Rather than working with the final action recognition classification output, as our aggregation methods do, the learned model works with the features provided by the pre-classification feature layer of the spatiotemporal action recognition network. Specifically, inspired by the concept of a multi-stage temporal convolutional network (MS-TCN) \cite{farha2019ms}, we construct dilated convolution models to integrate the local features from the classifier. Our model is a modification of the single-stage temporal convolutional network (SS-TCN), designed for action segmentation \cite{farha2019ms}, which itself is inspired by the WaveNet model \cite{oord_wavenet_2016} for raw audio waveform generation. 

The model takes a sequence of feature vectors $\vb{x}_0=(x_0^1,\ldots,x_0^T)$ of fixed length $T$, with each $x_0^t$ corresponding to a moment $t\in\{1,\ldots,T\}$ in time. (In our case, we have $T=575$, from sliding 2.5 s windows with stride 0.1 s across the 60 s videos, and obtaining feature vectors from our best action recognition model---see \secref{learning-based-segmentation} for details.) The sequence is fed through a 1D convolutional layer of kernel size 1 to reduce the channel size, and then through a number $L$ of successive 1D convolution layers $H_l$ of kernel size 3 for $l\in\{1,\ldots,L\}$, each producing a corresponding sequence of $T$ feature vectors $H_l(\vb{x}_{l-1})=\vb{x}_l=(x_l^1,\ldots,x_l^T)$ with the same channel size. The last sequence $\vb{x}_L$ of $T$ feature vectors is fed into a final 1D convolutional layer of kernel size 1, and then a softmax classification layer, to produce a sequence $\vb{y}=(y^1,\ldots,y^T)$ of class probabilities. The key feature of the model lies in the cascading dilation of its convolutional layers, depicted schematically \figref{Diagram}(a) and technically as follows:
\begin{equation*}
\label{eqn:network-dilation}
\adjustbox{scale=0.75,center}{%
\begin{tikzcd}[column sep = tiny,row sep = small]
& & & & \vdots & & & &\\
& & &  & x^t_l\arrow[u] &  & & &\\
x_{l-1}^{t-2^{l-1}}\arrow[urrrr] & & &  & x^t_{l-1}\arrow[u] &  & & & x_{l-1}^{t+2^{l-1}}\arrow[ullll]\\
& & & & \vdots\arrow[u] & & & & \\
& & &  & x^t_2\arrow[u] &  & & &\\
&  & x_1^{t-2}\arrow[urr] &  & x_1^t\arrow[u] &  & x_1^{t+2}\arrow[ull] &  &\\
& &  & x_0^{t-1}\arrow[ur] & x_0^t\arrow[u] & x_0^{t+1}\arrow[ul] &  & &\\
\end{tikzcd}}
\end{equation*}
Namely, while each convolutional layer $H_l$ has a kernel $k_l$ of fixed width of 3, the receptive field is essentially doubled at each layer, so for instance, $H_1$ acts locally by convolving $k_1\ast(x_0^{t-1}, x_0^t, x_0^{t+1})\mapsto x_1^t$,
$H_2$ acts locally by $k_2\ast(x_1^{t-2}, x_1^t, x_1^{t+2})\mapsto x_2^t$, and in general, $H_l$ acts locally via $k_l\ast(x_{l-1}^{t-2^{l-1}}, x_{l-1}^t, x_{l-1}^{t+2^{l-1}})\mapsto x_l^t$. (The kernels also act along the entire channel dimension, again, without modifying the channel size.) This dilated structure allows the model to exponentially grow its receptive field with the number of layers, at the cost of just linear parameter growth, enabling efficient processing of both short- and long-term dependencies.

The loss function 
\begin{equation}
L:=L_\text{class} + \lambda L_\text{smooth}
\end{equation}
combines a cross entropy loss with a smoothing loss via a scalar weight $\lambda$, chosen empirically. The standard cross entropy loss is defined by
\begin{equation}
L_\text{class} := \frac{1}{T}\sum_t -\log(y^t_{c_t}),
\end{equation}
where $y^t_c$ is the predicted probability at time $t$ for class $c$, and $c_t$ the ground truth class for time $t$. The smoothing loss is used in \cite{farha2019ms} to reduce rapid, unwarranted jumps in the segmentation assignments, and is defined as a truncated mean squared error between subsequent class log probabilities, 
\begin{equation}
L_\text{smooth}:=\frac{1}{TC}\sum_{t,c}\left(\lceil\log y^t_c - \log y^{t-1}_c\rceil^\kappa\right)^2,
\end{equation}
with $\lceil\cdot\rceil^\kappa$ denoting truncation at a threshold $\kappa$.

\tblresultCombo
\tblresultMultiThre

\section{Experimental Analysis}
\label{sec:experiments}

Here, we describe implementation details and experimental results for our non-nutritive sucking (NNS) action recognition and action segmentation models. For NNS action recognition, we test a range of convolutional and sequential neural backbones as well as the input modality (RGB vs optical flow), and also specifically gauge performance in challenging settings. For NNS action segmentation, we compare our fixed and learned methods for amalgamating the local analysis from our action recognition model into a global segmentation output, and also experiment with other backbones for local feature extraction, such as two-stream and 3D convolutional networks. 


\subsection{NNS Action Recognition Results}
\label{sec:nnsResults}
For the spatiotemporal core of our NNS action recognition, we experimented with four configurations of 2D convolutional networks, a 1-layer CNN, ResNet18, ResNet50, and ResNet101 \cite{he2016deep}; and three configurations of sequential networks, an LSTM, a bi-directional LSTM, and a transformer model \cite{vaswani2017attention}.
The models were trained for 50 epochs under a learning rate of 0.0001 using PyTorch 1.8.1 with CUDA 10.2, and the best model was chosen based on a held-out validation set.

We trained and tested this method with NNS clinical in-crib data from six infant subjects under a subject-wise leave-one-out cross-validation paradigm. Action recognition accuracies under are reported on the top left of \tblref{resultCombo}. Multiple thresholds are used to binarize the confidence scores while predicting to fully evaluate the pipeline. The results in \tblref{resultCombo} are from a confidence threshold of 0.8 
, and results under other thresholds are shown in \tblref{resultMultiThre}.

We elaborate on our choices for the convolutional and sequential networks, and their effect on the results:

\begin{description}
    \item[Convolutional] To explore the influence of the depth of CNN networks for spatial convolution, four CNN structures were utilized: a one-layer learnable convolution network to represent shallow CNN structure; the pre-trained ResNet18, ResNet50, and ResNet101 models for the middle to deep CNN structure. As the results are shown in \tblref{resultCombo}, all models with different CNNs were successfully learned and reached over $78.7\%$ accuracy on the NNS clinical in-crib dataset, which demonstrates the feasibility of the proposed CNN-LSTM model with optical flow input. The ResNet18-LSTM configuration performed best, achieving $95.8\%$ average accuracy over six infants using optical flow input. The strong performance (${\geq}78.1\%$) across all configurations indicates the viability of the overall method.
    \item[Sequential] We explore different structures of sequential dynamic event classifiers, including long short-term memory (LSTM), bi-directional LSTM, and transformer. The bi-directional has the same layer settings as the LSTM model, but the forward and backward outputs of the last node are concatenated before inputting into the fully connected layer. The transformer model is formed with 8 heads attention models and the feedforward network with 64 nodes. Bi-directional LSTM is the most robust one since it reaches the highest average accuracy over all CNN models both on the clinical in-crib dataset and on the in-the-wild dataset.  
\end{description}



\subsubsection{Evaluation In-the-Wild}
We also evaluated a model trained on all six infants from the NNS clinical in-crib dataset on the independent NNS in-the-wild dataset. Results on the bottom left of \tblref{resultCombo} again show strong cross-configuration performance (${\geq}78.1\%$), with ResNet101-Transformer reaching $92.3\%$, demonstrating strong generalizability of the method. As expected, models trained on the NNS clinical in-crib dataset tested worse on the independent NNS in-the-wild dataset. Interestingly, models with the smaller ResNet18 network suffered steep drop-offs in performance when tested on the in-the-wild data, while models based on the complex ResNet101 fared better under the domain shift. Beyond this, it is hard to identify clear trends between configurations or capacities and performance.

\subsubsection{Challenging Evaluation}
\label{sec:performance-challenging}

We explore the performance of our model under the difficult conditions present in the challenging subset of our NNS clinical in-crib dataset, which includes videos with infants in moving cribs, with faces partially occluded or under low light conditions. The top half of \tblref{challenge} shows performance of our action recognition model when tested on normal data, challenging data, and a mix of both, under the same subject-wise leave-one-out cross-validation configuration as before\footnote{For instance, within the cross-validation fold omitting R1, we train on the normal data from the other five subjects, and then evaluate on the normal R1 data, a mix of normal and challenging R1 data, and challenging R1 data, respectively.}. The performance on the challenging test data is particularly weak. For more context, we include precision and recall metrics as well as results under varying classifier confidence thresholds. These show that that model is indiscriminately sensitive, even at higher thresholds. 

Next, we experiment by including the challenging data in our training, again under the same subject-wise leave-one-out cross-validation configuration. The results are presented in the bottom half of \tblref{challenge}.  The performance is notably stronger on the challenging data, with higher thresholds yielding reasonably high precision as desired for some use cases, but overall performance is still below acceptability for most scientific purposes. Nonetheless, these tests suggest that more training with more challenging can help overcome issues arising from difficult conditions, and there is also room for specialized techniques to handle background movements, obstructions, and poor lighting.




\tblchallenge




\subsection{Action Segmentation Results} 
\label{sec:segmentation-results}

We evaluate both the fixed aggregation methods and our deep learning model for NNS action segmentation on the 60 s mixed-action videos in the NNS clinical in-crib dataset and the NNS in-the-wild dataset. For all methods, we use the standard evaluation metrics of average precision $\text{AP}_t$ and average recall $\text{AR}_t$ based on hits and misses defined by an intersection-over-union (IoU) with threshold $t$, across common thresholds $t\in\left\{0.1, 0.3, 0.5\right\}$\footnote{We follow definitions from \cite{idrees_thumos_2017}, with tiebreaks decided by IoU instead of confidence.}. Averages are taken with subjects given equal weight, and results are tabulated in \tblref{tblsegOld} for the aggregation-based method and \tblref{tblsegNew} for the learning-based model.
\tblsegOld

\subsubsection{Aggregation-Based Method}
We start with our best NNS action recognition model from \secref{nnsResults} (ResNet18-LSTM) as the local backbone, and test three aggregation-based methods for segmentation based on those local results. The test bed consists of our 60 s mixed activity clips, and we fellow the same leave-one-out cross-validation paradigm as we did for action recognition. In addition to the default classifier threshold of 0.5 used by our recognition model, we tested a 0.8 threshold to coax higher precision, as motivated in \secref{intro}. 
\segVis
The metrics in \tblref{tblsegOld} reveal strong performance from all methods and both confidence thresholds on both test sets. Generally, as expected, setting a higher confidence threshold or employing the more tempered tiled or smoothed aggregation methods favors precision, while lowering the confidence threshold or employing the more responsive sliding aggregation method favors recall. The results are excellent at the IoU threshold of 0.1 but degrade as the threshold is raised, suggesting that while these methods can readily perceive NNS behavior, they are still limited by the underlying ground truth annotator accuracy. The consistency of the performance of the model across both cross-validation testing in the clinical in-crib dataset and the independent testing on the NNS in-the-wild dataset suggests strong generalizability. \figref{segVis} visualizes predictions (and underlying confidence scores) of the sliding model configuration with a confidence threshold of 0.8, highlighting the excellent precision characteristics and illustrating the overall challenges of the detection problem. 

\tblsegNew
\subsubsection{Learning-Based Model}
\label{sec:learning-based-segmentation}
We use the same leave-one-out cross-validation pipeline to train and test for our learning-based model. However, rather than using final class predictions (NNS or non-NNS) from our NNS action recognition model, we work with the final pre-classification feature vectors. Specifically, working at the 10 Hz framerate, each 60 s video has 600 frames, and sliding 26 frame (2.5 s) windows across at a stride of 1 frame results in $T=575$ unique time points. For each window, we take $x_0^t$ to be the 128-dimensional pre-classification feature vector obtained by applying our ResNet18--LSTM model to that window. We use a dilated convolutional structure with $L=10$ layers, and loss weight $\lambda = 0.15$. The resulting performance metrics are tabulated in the bottom row of \tblref{tblsegNew}\footnote{The table also compares this pipeline with similar ones obtained by swapping our NNS action recognition model with other state-of-the-art action recognition models, trained on the same data, and again, with features taken from the pre-classification layer and fed into our segmentation model. We discuss these results in \secref{segmentation-ablation}.}

 
The results show that the learning-based model still can reach strong performance on both the clinical in-crib dataset and the in-the-wild dataset, attaining high precision as desired. Furthermore, compared to the aggregation-based methods (\tblref{tblsegOld}), the learned model exhibits more robust performance across multiple IoU thresholds while training and testing on the clinical in-crib dataset compared to the aggregation-based methods: the average precision ranges from $64.4\%$ to $88.4\%$ for the learning-based method, compared to $39.8\%$ to $93.5\%$ for the aggregation-based method. The learned model also achieves better precision and recall at higher IoU thresholds, suggesting that it provides more precise segments overall. 

\subsection{Comparison with the State-of-the-Art}

So far, we have tested various configurations of our NNS action recognition and NNS action segmentation pipelines, including different choices of architecture for deep network components. In this section, we instead test these pipelines against direct competitors: state-of-the-art action recognition and action segmentation models.

\subsubsection{Action Recognition Models}
\label{sec:other-methods}
Three widely recognized deep-learning-based action recognition methods are involved: I3D \cite{carreira2017quo}, X3D \cite{feichtenhofer2020x3d}, and 3D ResNet \cite{kataoka2020would}. Unlike the other two only using RGB input, the I3D method introduced another parallel network stream that takes optical flow as input and combines the RGB stream and optical flow stream together to make action prediction. Therefore, besides the original I3D two-stream structure,  we also performed fine-tuning on the RGB stream and optical flow stream independently to explore the effect of the input. The results are presented in \tblref{compareClassification}. As the results show, our proposed CNN-LSTM-based model reached the best performance on accuracy and precision for both the clinical in-crib dataset and in-the-wild dataset. Also, the I3D fine-tuned results align with the performance of our proposed method, which is optical flow input only has much better performance than the RGB input. The comparison shows the advantage of our model for dealing with subtle actions such as the NNS compared to the state-of-the-art models which are trained on general actions. 

\tblcompareClassification
\subsubsection{Action Segmentation Models}
\label{sec:other-segmentation-methods}
For the action segmentation models, we compare our deep-learning-based action segmentation model with the Global2Local \cite{gao2021global2local} method and ASFormer \cite{yi2021asformer}. All the models are trained and tested following the same pipeline as our proposed end-to-end-based method with the same feature input extracted from the pre-trained ResNet18-LSTM model. The comparisons are shown in \tblref{tblssegmentationCompare}, as the results show, our end-to-end-based method reached better average precision than the other methods under all IoU thresholds. Also, all models reached relatively close performance under all IoU thresholds with less than $15\%$ difference trained with the features extracted by the proposed pre-trained ResNet18-LSTM model, demonstrating our action recognition model feature extractor is general enough.

\figOpAblation
\tblssegmentationCompare

\subsection{Ablation Studies}

\subsubsection{Optical Flow Ablation}
Performance of all models with raw RGB input replacing optical flow frames can be found on the right side of \tblref{resultCombo}. 
The results are weak and close to random guessing, demonstrating the critical role played by optical flow in detecting the subtle NNS signal. This can also be seen clearly in the sample optical flow frames visualized in \figref{opFlow}.

We also evaluated multiple well-accepted optical flow methods including Farneback \cite{farneback2003two}, TV-L1 \cite{pock2007duality}, and RAFT \cite{teed2020raft}. The visualizations are shown in \figref{figOpAblation}. As the comparison shows, the accepted Croase2Fine method has the least background noise and strongest task-related area response.

\subsubsection{Feature Extractors}
\label{sec:segmentation-ablation}
We converted the fine-tuned I3D, X3D, and 3D ResNet models into feature extractors by removing the last layer and then substituted them for the feature extractor based on our NNS action recognition model, within our learning-based NNS action segmentation model. A comparison of performance results can be found in \tblref{tblsegNew}. Our specifically designed ResNet18-LSTM-based feature extractor performed better than all the other methods for all IoU thresholds in both datasets.


\section{Conclusion}
This article addresses the critical challenges surrounding infant feeding, where a delicate balance between sucking, swallowing, and breathing is required. Such challenges are especially pronounced in newborns and preterm infants, affecting approximately 2.8 million infants annually in the U.S. Traditional methods of assessing feeding readiness through subjective finger-in-mouth assessments of non-nutritive sucking (NNS) can pose discomfort and carry the risk of complications if inaccuracies occur. Our work introduces a pioneering approach to overcome these challenges by developing an automated, objective, video-based tool for tracking infant NNS. This tool not only has the potential to enhance real-time clinical decision-making but also holds promise for advancing research in infant neurodevelopmental diagnostics. Given the limited range of motor function and means of expression during infancy, NNS characteristics are invaluable indicators of neural and motor development. Furthermore, NNS has been proposed as a potential mechanism for reducing the risk of sudden infant death syndrome (SIDS), the leading cause of death among U.S. infants aged between 1 and 12 months. Our contributions include the creation of annotated infant video datasets, the development of an NNS classification system, an extensive comparison of spatiotemporal models, and the successful segmentation of NNS actions in longer video clips. These efforts lay the foundation for a fully automated computer vision assessment of NNS, enabling the extraction of critical sucking signal characteristics and contributing to our understanding of infant neurodevelopment and protective factors against SIDS.

\section{Acknowledgement} 
This research received support from MathWorks and the NSF-CAREER Grant \#2143882.

\bibliographystyle{model2-names.bst}\biboptions{authoryear}
\bibliography{paper}


\end{document}

%% file: figs.tex
\newcommand{\figPacDevice}{
\begin{figure}[t]
\centering
\includegraphics[width=\linewidth]{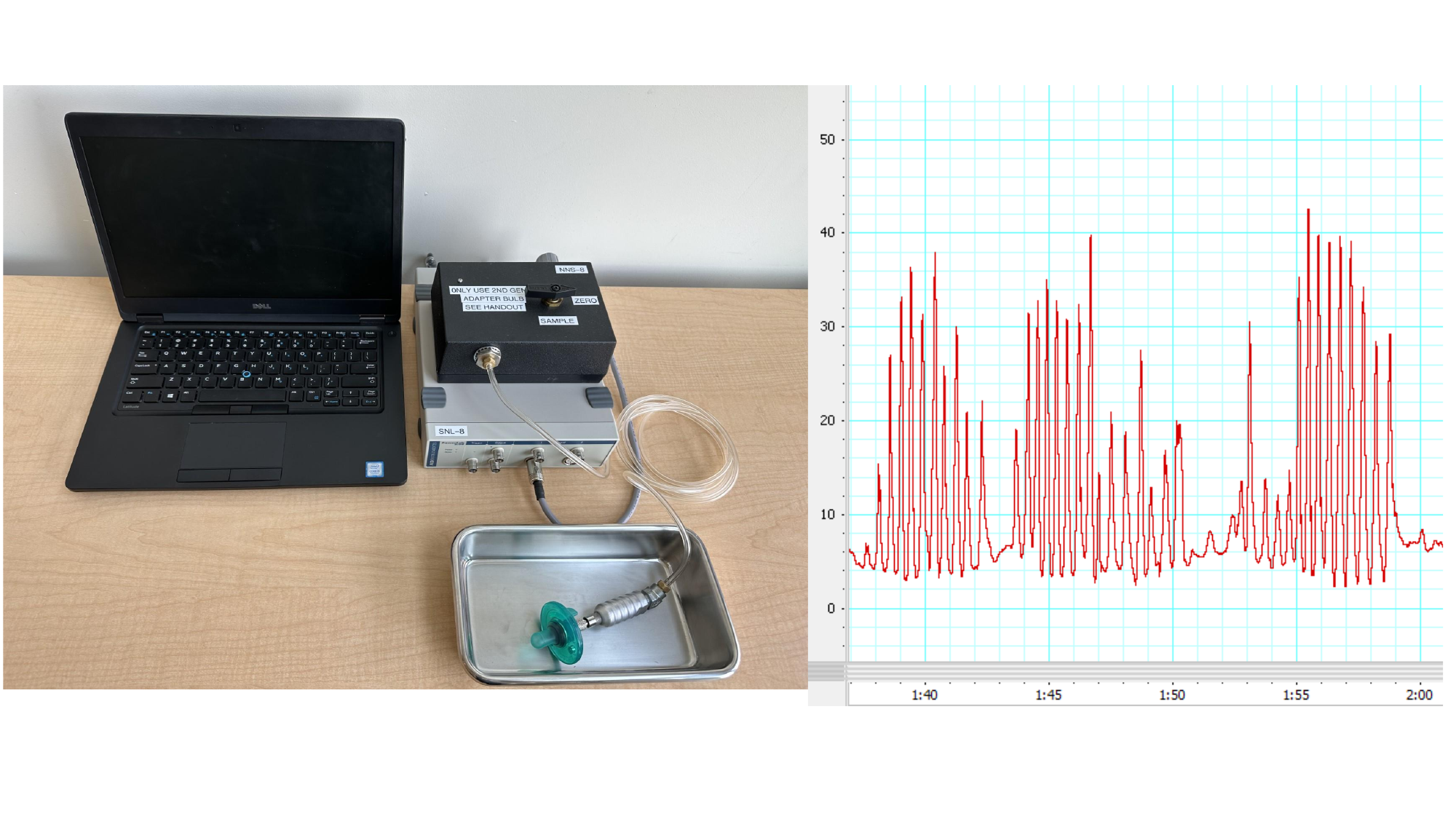}
 \caption{A pressure transducer pacifier device \textit{(left)} and an extracted non-nutritive sucking (NNS) signal obtained from the device \textit{(right)} \cite{martens_changes_2020}. While such a tool can provide reliable, high resolution measurements, it is expensive and limited to research use, and could interfere with the natural sucking behavior. Our computer vision method based on spatiotemporal neural networks enables completely contactless detection and segmentation of NNS activity.}
\label{fig:figPacDevice}
\end{figure}
}

\newcommand{\figGallery}{
\begin{figure*}[t]
\centering
\includegraphics[width=\textwidth]{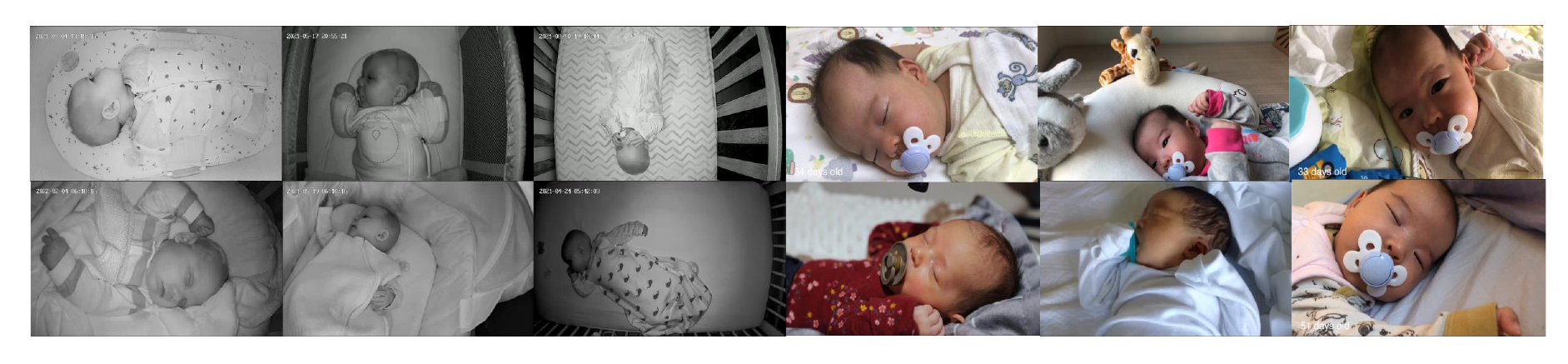}
 \caption{\textit{Left:} Still frames extracted from our NNS clinical in-crib dataset, consisting of 183 hours of nigh-time in-crib baby monitor footage from 19 infants. \textit{Right:} Frames from our our publicly available NNS in-the-wild dataset, drawn from public sources to complement our clinical dataset with further variety. Both datasets feature timestamp annotations drawn from behavioral coding for NNS activity and pacifier use.}
\label{fig:nnsSignal}
\end{figure*}
}

\newcommand{\figDiagram}{
\begin{figure*}[t]
\centering
\includegraphics[width=\linewidth]{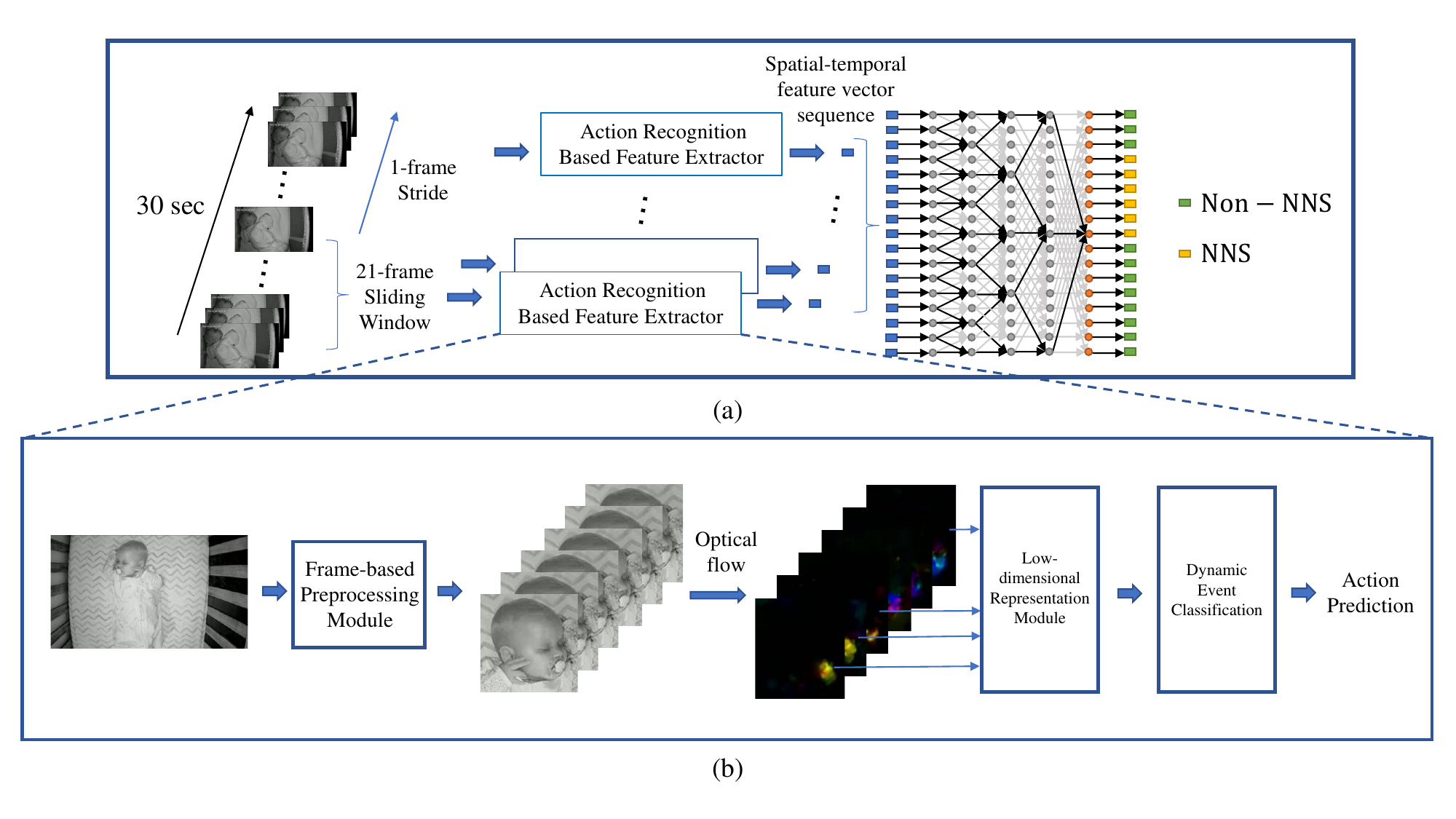}
\caption{ 
\textit{(a):} Our NNS segmentation pipeline, based on aggregating local results of NNS action recognition in sliding windows. Features for each sliding window are extracted using the proposed action-recognition-based feature extractor and then input into the dilated convolution network for frame-based action prediction and achieve action segmentation.
\textit{(b):} Our NNS action recognition pipeline, which applies dense optical flow to preprocessed frames and passes features through a convolutional layer followed by a temporal layer to obtain an action prediction based on spatiotemporal information.}
\label{fig:Diagram}
\end{figure*}
}

\newcommand{\setup}{
\begin{figure}[ht]
\centering
\includegraphics[width=\linewidth]{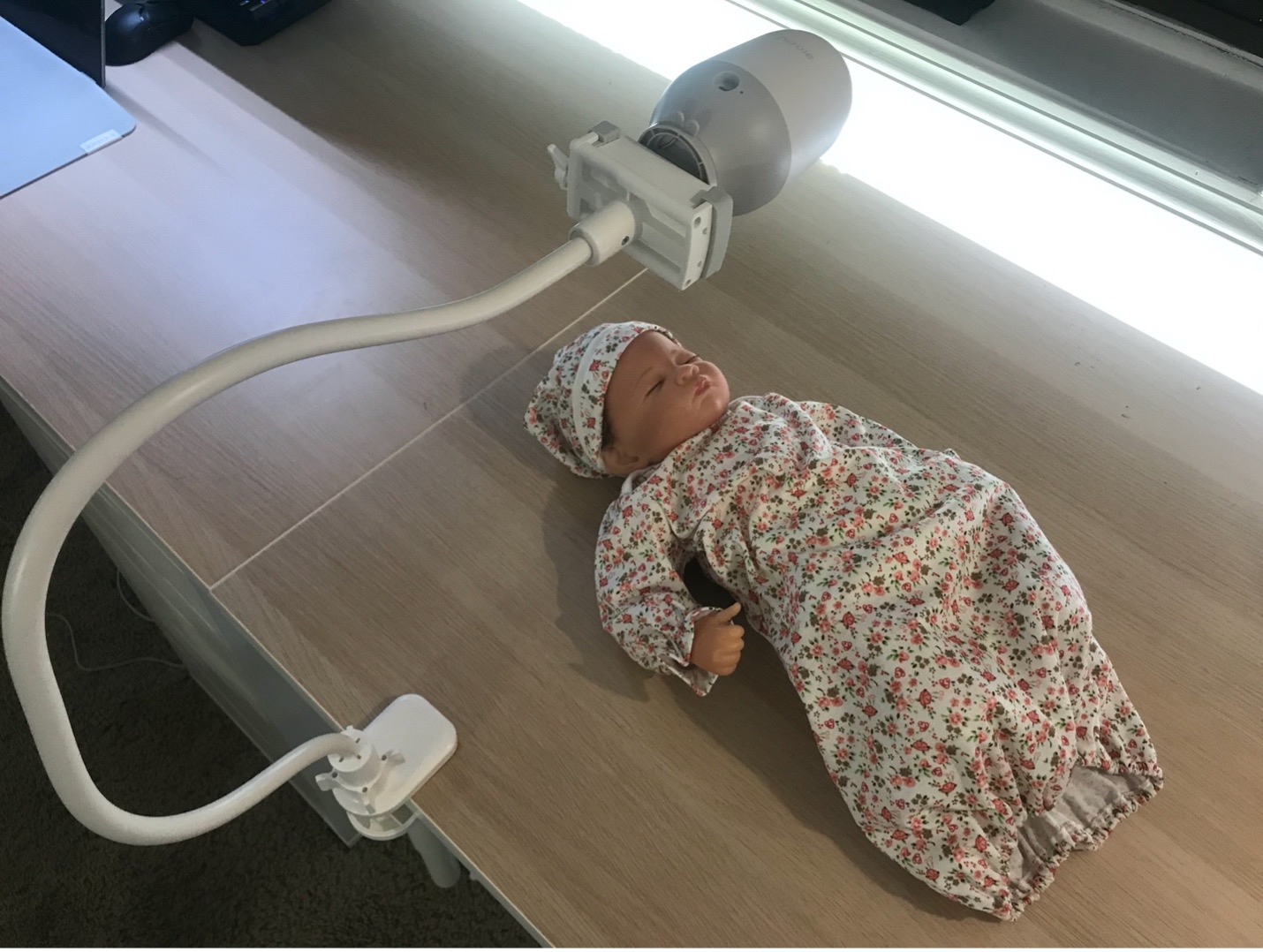}
\caption{Suggested baby monitor placement for study participants for our NNS clinical in-crib dataset. Videos were shot by parents or caregivers in 2021 and 2022 during the pandemic. They are long and feature a wide variety of natural infant behavior, including napping, sleeping, tossing and turning, crying, and caregiver interactions such as pacifier insertion, patting, removal from the crib, and more, yielding a true-to-life but technically challenging data source.}
\label{fig:setup}
\end{figure}
}

\newcommand{\annotSoft}{
\begin{figure}[ht]
\centering
\includegraphics[width=\linewidth]{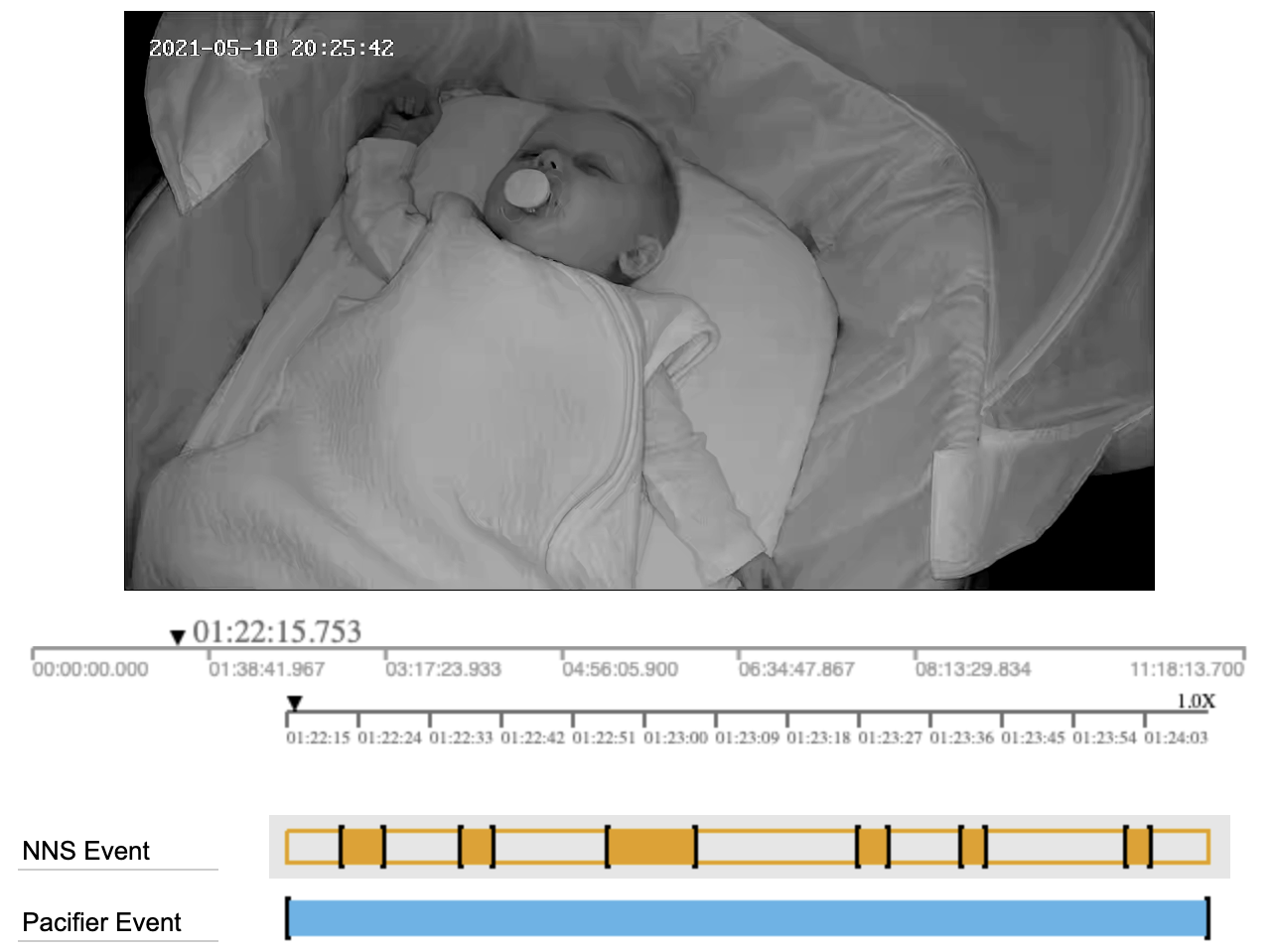}
\caption{NNS annotation tool \cite{dutta2019vgg} used by our behavioral coding team specifically trained for this task. For the NNS clinical in-crib dataset annotations, duplicate coding and systematic checks were implemented to ensure reliability over the hundreds of hours of footage.}
\label{fig:annotSoft}
\end{figure}
}

\newcommand{\opFlow}{
\begin{figure*}[ht]
\centering
\includegraphics[width=\linewidth]{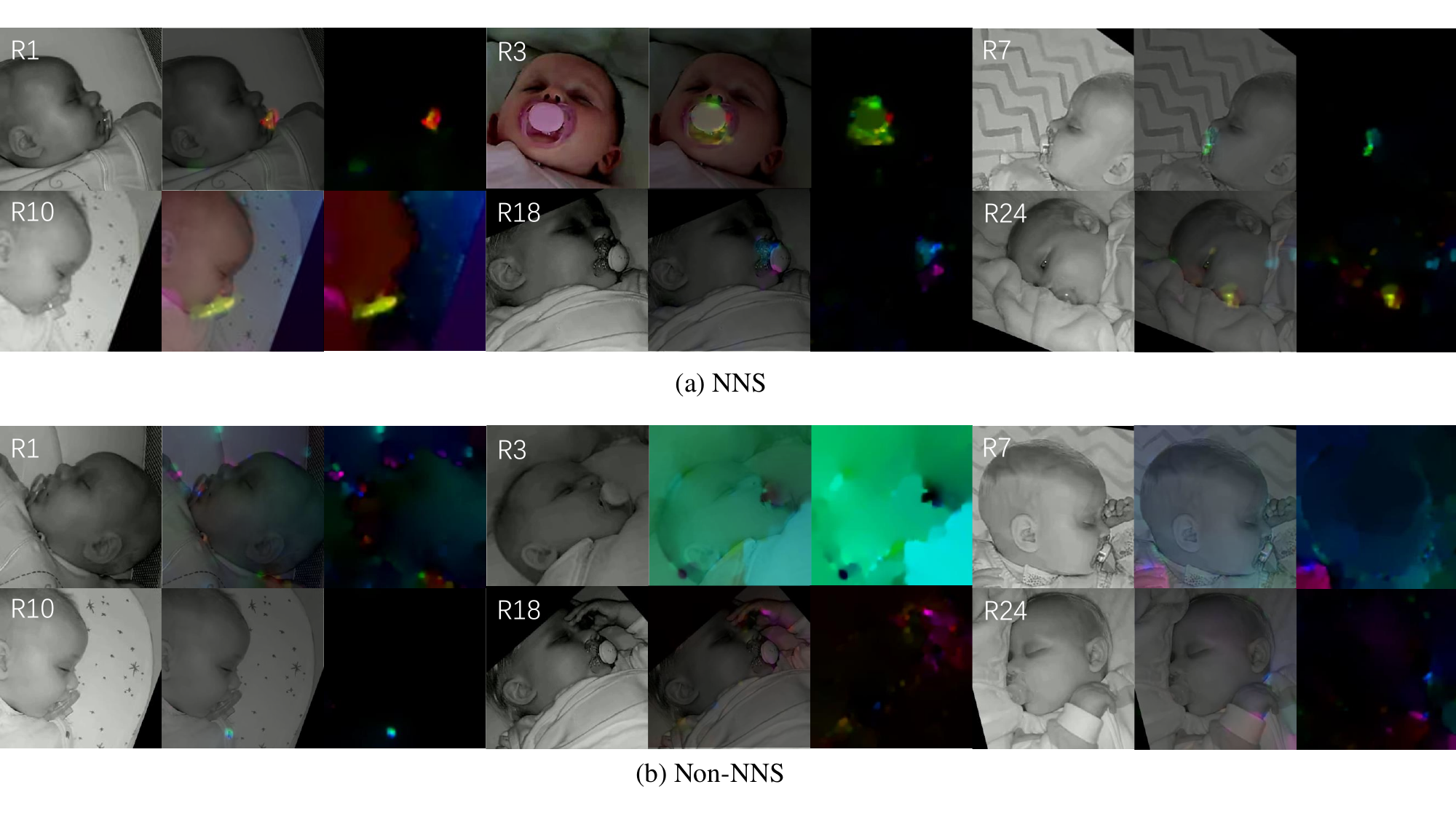}
\caption{Visualizations of the optical flow processing on our NNS clinical in-crib datasets. For each subject (R\#), we show the video frame on the left, a color representation of the optical flow field on the right, and the superposition of the two in the middle. Examples of NNS action in-progress are given on top in \textit{(a)}, and examples of no NNS action taking place are given below in \textit{(b)}, illustrating the effectiveness of optical flow in discerning the subtle sucking signal. Optical flow can be noisy, for instance, picking up on video encoding artefacts in R3(b). See the optical flow ablation study in \secref{nnsResults} for corresponding numerical results.}
\label{fig:opFlow}
\end{figure*}
}

\newcommand{\figOpAblation}{
\begin{figure}[ht]
\centering
\includegraphics[width=\linewidth]{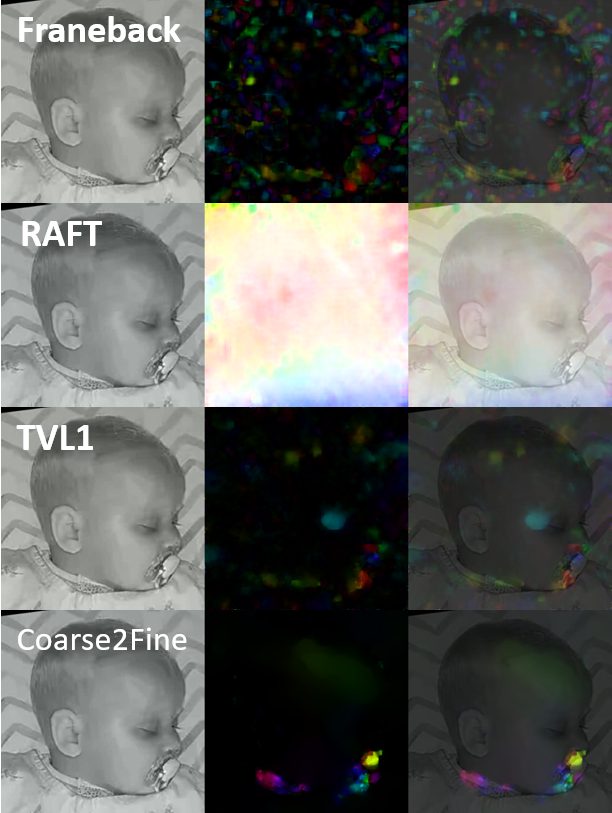}
\caption{Comparison of optical flow results from four widely used algorithms on a video from our NNS clinical in-crib dataset, illustrating the ability of Coarse2Fine to most cleanly isolate the NNS movement from background noise. For each method, the video frame is shown on the left, the optical flow field in the middle, and the superposition of the two on the right.}
\vspace{-.5cm}
\label{fig:figOpAblation}
\end{figure}
}

\newcommand{\segVis}{
\begin{figure*}[ht]
\centering
\includegraphics[width=\linewidth]{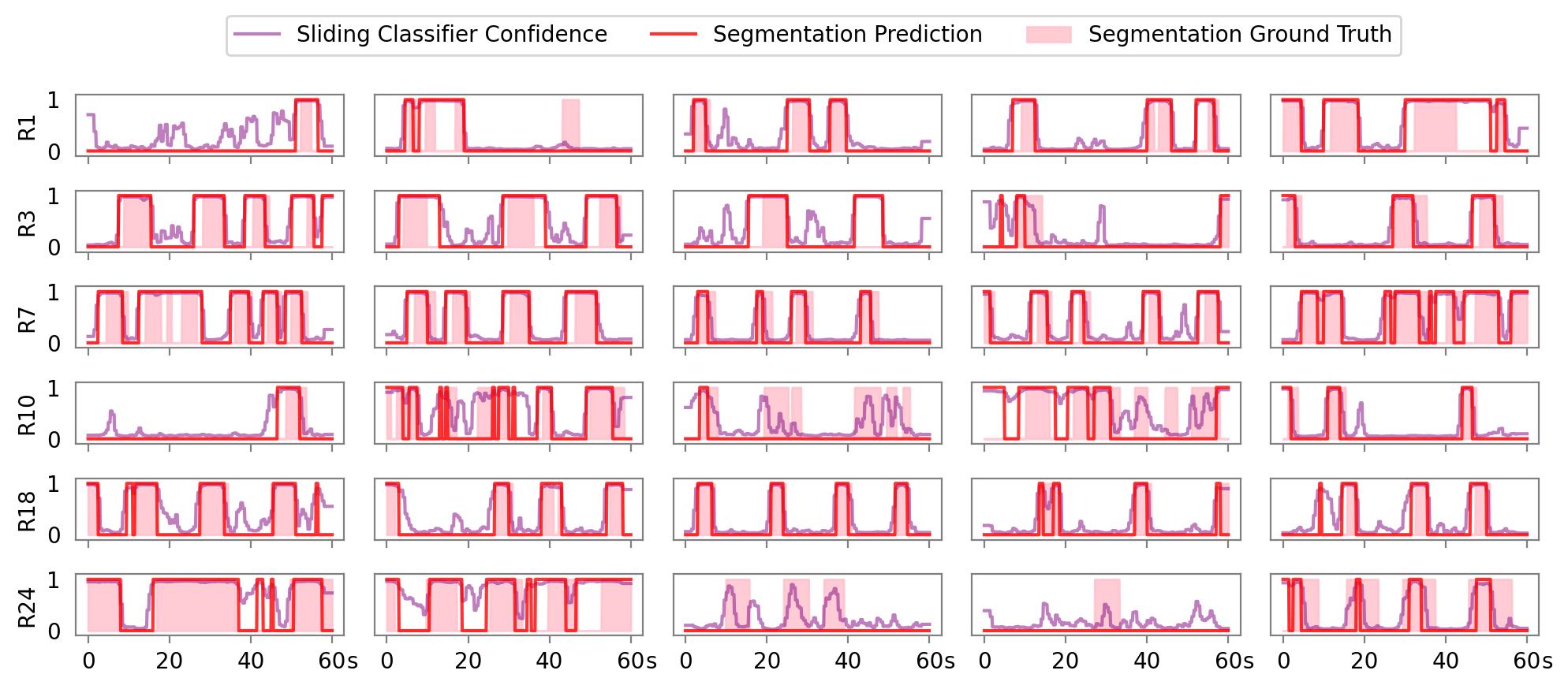}

\caption{Segmentation predictions and ground truth for each 60 s mixed clip from the NNS clinical in-bed dataset, under the sliding window aggregation model configuration and with a confidence threshold of 0.9, boosting precision at the cost of recall.}
\label{fig:segVis}
\end{figure*}
}

%% file: tabs.tex
\newcommand{\tblcohenkappa}{

\begin{table}[t]

\ra{1.0} 
\centering 
\caption{Cohen $\kappa$ inter-rater reliability scores for behavioral coding between pairs of annotators, for NNS activity, pacifier use, and NNS activity during pacifier usage, in our NNS clinical in-crib dataset. The $\kappa$ scores are computed after converting start and end timestamp data to binary time series based on incidence within uniform windows of the specified lengths.}
\label{tbl:cohen-kappa}
\resizebox{\columnwidth}{!}{
\begin{tabular}{@{}lrrlr@{}}
    \toprule
    Event & Window (s) & Mean $\kappa$ & Interpretation & SD $\kappa$\\
    \toprule
    NNS (all) & 0.1 & 58.0 & Weak & 5.2\\
    NNS during pacifier usage & 0.1 & 69.2 & Moderate & 7.8 \\
    Pacifier usage & 0.1 & 98.0 & Almost perfect & 2.9\\
    \midrule
    NNS (all) & 10 & 66.4 & Moderate & 4.4\\
    NNS during pacifier usage & 10 & 82.8 & Strong & 9.5\\
    Pacifier usage & 10 & 98.1 & Almost perfect & 2.9\\
    \bottomrule
\end{tabular}}
\end{table}
}

\newcommand{\tblnnsstats}{
\begin{table*}[ht]
\caption{Biographical data and NNS and pacifier event statistics for the 10 pacifier-using infants from our NNS clinical in-crib study, six of whom ($\star$) engaged in enough NNS activity for use in machine learning. \textit{Age} is at time of video capture, \textit{BGA} the birth gestational age, \textit{BWt} the birth weight, \textit{Dur} the cumulative event duration, \textit{C-$\kappa$} the Cohen $\kappa$ annotator agreement (incidence on 10 s windows), \textit{Ct} the count, and \textit{Len} the length of individual events. Biographical data are self-reported (hence whole numbers), and event data are averaged from the two annotators (hence fractional counts).}
\label{tbl:nns-stats}
\centering
\ra{1.1} 
\resizebox{\textwidth}{!}{\begin{tabular}{@{}lrrrrrrrrrrrrrrrrrrrrr@{}}
    \toprule
    & & \multicolumn{4}{c}{Biographical Data} & & Vid & & \multicolumn{6}{c}{\text{NNS Events (During Pacifier Use)}} & \phantom{a} & \multicolumn{6}{c}{\text{Pacifier Events}}\\ 
    \cmidrule[0.5pt]{3-6} \cmidrule[0.5pt]{8-8} \cmidrule[0.5pt]{10-15} \cmidrule[0.5pt]{17-22}
    Sbj & & Sex & Age & BGA & BWt && Dur && C-$\kappa$ & Ct & Ct/h & Dur & Dur/h & Len & & C-$\kappa$ & Ct & Ct/h & Dur & Dur/h & Len\\
     &&& d & wk & oz && h &&& \# & \#/h & min & min/h & s & &  & \# & \#/h & min & min/h & min\\
    \midrule
    R1$\star$ && M & 100 & 40 & 130 && 10.9 && 0.74 & 636.0 & 58.4 & 39.5 & 3.6 & 3.9 && 0.92 & 13.0 & 1.2 & 219.5 & 20.1 & 16.9\\
    R3$\star$ && F & 98  & 39 & 106 && 11.3 && 0.69 & 490.5 & 43.4 & 51.2 & 4.5 & 6.3 && 0.98 & 10.0 & 0.9 & 270.0 & 23.9 & 27.3\\
    R7$\star$ && F & 103 & 39 & 109 && 13.5 && 0.91 & 817.0 & 60.4 & 60.7 & 4.5 & 4.4 && 1.00 & 5.0 & 0.4 & 214.1 & 15.8 & 44.6\\
    R9         && M & 82  & 40 & 145 && 3.1 && 0.83 & 18.0 & 5.8 & 1.8 & 0.6 & 6.4 && 1.00 & 9.0 & 2.9 & 4.5 & 1.4 & 0.5\\
    R10$\star$ && F & 114 & 39 & 121 && 5.0 && 0.90 & 92.5 & 18.6 & 5.2 & 1.1 & 3.6 && 1.00 & 7.0 & 1.4 & 28.7 & 5.8 & 4.1\\
    R12        && F & 112 & 41 & 101 && 13.1 && 0.96 & 79.5 & 6.1 & 6.3 & 0.5 & 4.8 && 0.99 & 3.5 & 0.3 & 14.3 & 1.1 & 4.2\\
    R15        && F & 102 & 40 & 110 && 14.7 && 0.86 & 106.0 & 7.2 & 14.1 & 1.0 & 8.1 && 0.99 & 7.0 & 0.5 & 52.1 & 3.6 & 8.1\\
    R18$\star$ && F & 142 & 37 & 99  && 6.3 && 0.84 & 115.0 & 18.2 & 6.4 & 1.0 & 3.3 && 0.99 & 4.0 & 0.6 & 39.3 & 6.2 & 10.5\\
    R23        && F & 151 & 42 & 106 && 12.8 && 0.79 & 30.0 & 2.3 & 1.6 & 0.1 & 3.2 && 0.94 & 7.5 & 0.6 & 8.9 & 0.7 & 1.2\\
    R24$\star$ && M & 120 & 39 & 129 && 10.7 && 0.80 & 527.5 & 49.4 & 67.2 & 6.3 & 8.1 && 0.97 & 6.5 & 0.6 & 232.3 & 21.7 & 35.9\\
    \midrule
    Mean       && -  & 112.4 & 39.8 & 115.6 && 10.1 && 0.83 & 291.2 & 27.0 & 25.4 & 2.3 & 5.2 && 0.98 & 7.2 & 0.9 & 108.4 & 10.0 & 15.3 \\
    Std       && -  & 20.8  & 1.4  & 15.0  && 3.9 && 0.08 & 295.0 & 23.4 & 26.3 & 2.2 & 1.9 && 0.03 & 2.9 & 0.8 & 110.0 & 9.3 & 15.5\\
    \midrule
    Mean$\star$ && - & 112.8 & 38.9 & 115.7 && 9.6 && 0.81 & 446.4 & 41.4 & 38.4 & 3.5 & 4.9 && 0.98 & 7.6 & 0.8 & 167.3 & 15.6 & 23.3 \\
    Std$\star$ && - & 16.4  & 1.0  & 12.9  && 3.3 && 0.09 & 288.8 & 18.9 & 26.9 & 2.1 & 1.9 && 0.03 & 3.4 & 0.4 & 105.1 & 7.9 & 15.5\\
    \bottomrule
\end{tabular}}
\end{table*}
}

\newcommand{\tblresultCombo}{
\begin{table*}[htb!]
  \caption{Classification accuracy of our NNS action recognition model, under various convolutional and temporal configurations and two image modalities. We test on the NNS clinical in-crib data under subject-wise leave-one-out cross-validation, and on the NNS in-the-wild data directly, both with balanced classes. The strongest results are in bold. The results reported in the current table are under $0.8$ confidence threshold, the rest results under other thresholds are shown in the supplementary material.}
    \centering
    \ra{1.0} 
    \resizebox{\textwidth}{!}{\begin{tabular}{@{}llrrrrrrrrrrr@{}}
        \toprule
        & \phantom{a} & \phantom{a} & \multicolumn{4}{c}{\text{Optical Flow}} & \phantom{a} & \multicolumn{4}{c}{\text{RGB}}\\ 
        \cmidrule[0.5pt]{4-7} \cmidrule[0.5pt]{9-12}
        & \diagbox{Sequential}{Convolutional} &  & 1-lr. CNN & ResNet18 & ResNet50 & ResNet101 && 1-lr. CNN & ResNet18 & ResNet50 & ResNet101\\
        Dataset & & \# Tr. Params. & 333K & 154K & 614K & 614K && 333K & 154K & 614K & 614K \\
        \midrule
        \multirow{3}*{Clinical} 
          & Transformer & 393K & 79.2 & 92.5 & 94.0 & 78.1 &                    & 50.3 & 50.3 & 50.5 & 50.0\\
          ~ & LSTM        &  418K & 85.8 & \textbf{95.8} & 82.3 & 85.2 &  & 51.2 & 50.2 & 50.0 & 50.0\\
		~ & Bi-LSTM     & 535K  & 78.6 & 93.4 & 90.6 & 85.8 &           & \textbf{52.3} & 49.8 & 50.0 & 50.0\\

        \midrule
        \multirow{3}*{In-the-wild} 
        & Transformer & 393K  & 81.5 & 81.2 & 84.0 & \textbf{94.6} &  & \textbf{56.8} & 45.9 & 45.9 & 45.9\\
        ~  & LSTM        & 418K   & 86.3 & 78.4 & 86.0 & 78.7 &       & 52.0 & 45.9 & 50.2 & 50.2\\
		~ & Bi-LSTM     & 535K  & 78.1 & 87.1 & 86.5 & 86.3 &           & 54.4 & 51.7 & 50.2 & 49.8\\
        
        \bottomrule
    \end{tabular}}
    \label{tbl:resultCombo}
\end{table*}
}

\newcommand{\tblchallenge}{
\begin{table*}[t]
\ra{0.9} 
\centering 
\caption{ Classification performance of our best LSTM-ResNet18 model and tested on different mixes of the NNS clinical in-crib normal and challenging subsets under different classification thresholds. The upper half is the evaluation using the model trained on normal data only, and the lower half is the evaluation using the model trained on normal and challenging data mixture. The results are the averaged classification accuracy, precision, and recall evaluated on the six subjects in the in-crib dataset. The strongest results are in bold for both training sets up.}
\label{tbl:challenge}
\resizebox{0.75\textwidth}{!}{
\begin{tabular}{@{}lrrrrrrrrrrrrr@{}}
    \toprule
    \diagbox{Training}{Testing}  && \multicolumn{3}{c}{Normal Only} && \multicolumn{3}{c}{Challenge Only} && \multicolumn{3}{c}{Normal + Challenge}\\
    \cmidrule[0.5pt]{3-5} \cmidrule[0.5pt]{7-9} \cmidrule[0.5pt]{11-13}
    &  Thres. & Acc. & Prec. & Rec. && Acc. & Prec. & Rec. && Acc. & Prec. & Rec.\\
    \midrule
    \multirow{5}*{Normal Only} 
    & 0.5  & 94.9 & 91.1 & \textbf{95.3} && 52.9 & 51.6 & \textbf{99.2} && 84.9 & 78.6 & 96.0\\
    ~& 0.6  & 94.4 & 92.7 & 93.6 && 52.9 & 51.6 & \textbf{}{99.2} && 84.6 & 79.3 & 94.7\\
    ~& 0.7  & 94.9 & 94.2 & 91.9 && 52.9 & 51.6 & 98.3 && \textbf{85.0} & 80.1 & 93.2\\
    ~& 0.8  & \textbf{95.8} & 94.8 & 93.4 && 54.6 & 52.4 & 97.5 && \textbf{85.0} & 80.7 & 91.0\\
    ~& 0.9  & 92.2 & \textbf{97.5} & 87.7 && \textbf{55.0} & \textbf{53.0} & 87.5 && 82.8 & \textbf{82.8} & \textbf{82.8}\\

    \midrule
    \multirow{5}*{\makecell[l]{Normal + Challenge}} 
    & 0.5  & 90.6 & 87.2 & \textbf{96.9} && 58.8 & 55.7 & \textbf{95.8} && 84.5 & 78.3 & \textbf{96.7}\\
    ~& 0.6 & 91.7 & 89.0 & 96.3 && 60.4 & 56.9 & 93.3 && 85.4 & 80.1 & 95.7\\
    ~& 0.7 & 91.4 & 89.8 & 94.4 && 60.0 & 59.0 & 83.3 && 85.1 & 81.7 & 92.2\\
    ~& 0.8 & \textbf{92.2} & 93.2 & 91.9 && \textbf{62.5} & 64.8 & 73.3 && \textbf{86.3} & 86.1 & 88.2\\
    ~& 0.9 & 83.6 & \textbf{97.0} & 83.8 && 60.0 & \textbf{73.3} & 49.0 && 78.9 & \textbf{92.7} & 76.8\\ 
    \bottomrule
\end{tabular}}
\end{table*}
}

\newcommand{\tblsegOld}{
\begin{table*}[htb!]
  \caption{Average precision $\text{AP}_t$ and average recall $\text{AR}_t$ performance for various IoU thresholds $t$ of our NNS segmentation model. We test three local classification aggregation methods and two different classifier confidence thresholds. Precision-recall pairs with the highest precision in each threshold configuration are in bold.}
    \centering
    \ra{1.0} 
    \resizebox{\textwidth}{!}{\begin{tabular}{@{}llrrrrrrrrrrrrrrr@{}}
        \toprule
        & \phantom{a} & & \multicolumn{6}{c}{\text{Classifier Confidence Threshold = 0.8}} & \phantom{a} & \multicolumn{6}{c}{\text{Classifier Confidence Threshold = 0.5}}\\ 
        \cmidrule[0.5pt]{4-9} \cmidrule[0.5pt]{11-16}
        Dataset & Method && $\text{AP}_{0.1}$ & $\text{AR}_{0.1}$ & $\text{AP}_{0.3}$ & $\text{AR}_{0.3}$ & $\text{AP}_{0.5}$ & $\text{AR}_{0.5}$ & & $\text{AP}_{0.1}$ & $\text{AR}_{0.1}$ & $\text{AP}_{0.3}$ & $\text{AR}_{0.3}$ & $\text{AP}_{0.5}$ & $\text{AR}_{0.5}$ \\
        
        \midrule
        \multirow{3}*{\makecell[l]{Clinical}}   
        & Tiled    && \textbf{93.5} &  \textbf{92.9} & \textbf{75.7} & \textbf{76.9} & \textbf{39.8} & 40.4 && \textbf{90.3} &        \textbf{91.5} &       \textbf{77.8} &        \textbf{76.6} &       \textbf{51.0} &        \textbf{50.8}\\
        & Sliding  && 76.5 & 90.1 & 63.5 & 76.4 & 36.1 & \textbf{43.4} && 78.3 &        92.7 &       70.3 &        82.5 &       45.4 &        53.1 \\
        & Smoothed && 90.2 & 79.9 & 75.6 & 65.9 & 33.5 & 30.8 && 86.9 & 91.0 & 74.0 & 72.9 & 42.6 & 44.8  \\
	    \midrule
        \multirow{3}*{\makecell[l]{In-the-\\wild}}  
        &  Tiled  && \textbf{96.0} &  \textbf{90.4} & \textbf{77.7} & \textbf{74.8} & \textbf{67.6} & 63.4 && \textbf{90.8} &        \textbf{84.2} &       \textbf{80.5} &        \textbf{74.4} &       67.9 &        63.5 \\
        & Sliding  && 84.9 & 87.4 & 66.0 & 72.4 & 61.9 & \textbf{66.1} && 79.0 &        85.1 &       67.2 &        72.7 &       62.8 &        66.5 \\
        & Smoothed  && 94.3 & 80.3 & 73.7 & 65.9 & 62.0 & 55.0 && 90.0 &        78.7 &       77.0 &        67.5 &       \textbf{72.2} &        \textbf{62.6}\\
        \bottomrule
    \end{tabular}}
    \label{tbl:tblsegOld}
\end{table*}
}

\newcommand{\tblsegNew}{
\begin{table}[htb!]
  \caption{The action segmentation performance of our proposed deep-learning-based model. Other state-of-the-art action recognition models including I3D \cite{carreira2017quo}, X3D \cite{feichtenhofer2020x3d}, and 3D ResNet \cite{kataoka2020would} are converted into feature extractors and follow the same pipeline to input into our proposed model.}
    \centering
    \ra{1.0} 
    \resizebox{\columnwidth}{!}{\begin{tabular}{@{}llrrrrrr@{}}
        \toprule
        Dataset & Method & $\text{AP}_{0.1}$ & $\text{AR}_{0.1}$ & $\text{AP}_{0.3}$ & $\text{AR}_{0.3}$ & $\text{AP}_{0.5}$ & $\text{AR}_{0.5}$\\
        
        \midrule
        \multirow{3}*{\makecell{Clinical}}   & I3D & 50.7 & 63.4 & 37.6 & 44.1 & 17.8 & 20.8\\
                                                & X3D & 45.4 & 54.8 & 26.4 & 30.9 & 9.6 & 12.5\\
                                                & 3D ResNet & 35.8 & 13.5 & 26.4 & 10.3 & 19.4 & 8.6\\
                                                & Ours & \textbf{88.4} & \textbf{86.5} & \textbf{80.5} & \textbf{76.7} & \textbf{64.4} & \textbf{63.5}\\
	    \midrule
        \multirow{3}*{\makecell{In-the-\\wild}}  & I3D  & 75.4 & 81.0 & 59.2 & 62.7 & 32.2 & 35.7\\
                                                & X3D & 68.8 & 44.2 & 34.3 & 26.1 & 18.6 & 18.6\\
                                                & 3D ResNet & 62.5 & 51.2 & 30.3 & 28.0 & 18.4 & 15.3\\
                                                & Ours & \textbf{91.0} & \textbf{88.5} & \textbf{78.3} & \textbf{74.1} & \textbf{58.6} & \textbf{54.7}\\
        \bottomrule
    \end{tabular}}
    \label{tbl:tblsegNew}
\end{table}
}

\newcommand{\tblcompareClassification}{
\begin{table}[htb!]
  \caption{Comparison with the state-of-the-art action recognition methods. I3D \cite{carreira2017quo} represents the original two-stream (RGB + optical flow) structure. I3D RGB and I3D OP represent the cases in which only the RGB stream or optical flow stream of the I3D pre-trained model is used. X3D \cite{feichtenhofer2020x3d}, and 3D ResNet \cite{kataoka2020would} models are fine-tuned on the clinical in-crib dataset and tested on the in-the-wild dataset. The strongest results are in bold for both datasets.}
    \centering
    \ra{1.0} 
    \resizebox{\columnwidth}{!}{\begin{tabular}{@{}llcccccc@{}}
        \toprule
        Data & Evaluation & I3D & I3D RGB & I3D OP & X3D & 3D ResNet & Ours\\
        \midrule
        \multirow{3}*{\makecell{Clinical}} 
          & Acc & 77.4 & 67.4 & 80.5 & 74.5 & 72.9 & \textbf{95.8}\\
        ~ & Precision & 81.0 & 73.8 & 83.1 & 85.0 & 81.9 & \textbf{94.3}\\
	   ~ & Recall & 60.6 & 67.2 & 77.9 & 66.6 & 65.0 & \textbf{92.7}\\
        \midrule
        \multirow{3}*{\makecell{In-the-Wild}} 
          & Acc & 65.0 & 65.9 & 69.7 & 77.5 & 73.7 & \textbf{78.4}\\
        ~ & Precision & 60.6 & 73.0 & 65.1 & 80.6 & 82.5 & \textbf{83.8}\\
	~ & Recall & \textbf{100.0} & 82.6 & \textbf{100.0} & 84.0 & 80.2 & 81.5\\

        \bottomrule
    \end{tabular}}
    \label{tbl:compareClassification}
\end{table}
}

\newcommand{\tblssegmentationCompare}{
\begin{table}[htb!]
  \caption{Comparison with the state-of-the-art action segmentation methods including Global2Local \cite{gao2021global2local} method and ASFormer \cite{yi2021asformer}. 
  Average precision $\text{AP}_t$ and average recall $\text{AR}_t$ performance for various IoU thresholds $t$ of each model.}
    \centering
    \ra{1.0} 
    \resizebox{\columnwidth}{!}{\begin{tabular}{@{}llrrrrrr@{}}
        \toprule
        Dataset & Method & $\text{AP}_{0.1}$ & $\text{AR}_{0.1}$ & $\text{AP}_{0.3}$ & $\text{AR}_{0.3}$ & $\text{AP}_{0.5}$ & $\text{AR}_{0.5}$\\
        
        \midrule
        \multirow{3}*{\makecell{Clinical}}   & Global2Local & 79.7 & \textbf{89.1} & 72.7 & 76.7 & 53.8 & 59.0\\
                                                & ASFormer & 85.8 & 84.4 & 75.7 & 73.1 & 61.6 & 60.2\\
                                                & Ours & \textbf{88.4} & 86.5 & \textbf{80.5} & \textbf{80.2} & \textbf{64.4} & \textbf{63.5}\\
	    \midrule
        \multirow{3}*{\makecell{In-the-wild}}  & Global2Local & 83.6 & 86.4 & 78.3 & 74.1 & 48.4 & 49.9\\
                                                & ASFormer & 90.8 & 87.9 & 77.5 & 72.6 & 58.6 & 54.7\\
                                                & Ours & \textbf{91.0} & \textbf{88.5} & \textbf{78.8} & \textbf{78.6} &  \textbf{63.3} & \textbf{60.5}\\
        \bottomrule
    \end{tabular}}
    \label{tbl:tblssegmentationCompare}
\end{table}
}

\newcommand{\tblresultMultiThre}{
\begin{table*}[htb!]
  \caption{Classification accuracy of our NNS action recognition model, under various convolutional and temporal configurations and two image modalities. We test on the NNS clinical in-crib data under subject-wise leave-one-out cross-validation, and on the NNS in-the-wild data directly, both with balanced classes. The strongest results are in bold. }
    \centering
    \ra{1} 
    \resizebox{\textwidth}{!}{\begin{tabular}{@{}lllrrrrrrrrr@{}}
        \toprule
        & &  & \multicolumn{4}{c}{\text{Optical Flow}} &  & \multicolumn{4}{c}{\text{RGB}}\\ 
        \cmidrule[0.5pt]{4-7} \cmidrule[0.5pt]{9-12}
        Threshold & Dataset & Sequential &  1-lr. CNN & ResNet18 & ResNet50 & ResNet101 && 1-lr. CNN & ResNet18 & ResNet50 & ResNet101\\

        \midrule
        \multirow{6}*{0.5} 
        & \multirow{3}*{Clinical} 
          & Transformer  & 90.9 & 89.4 & 88.5 & 89.2 && 63.5 & 53.5 & 56.4 & 47.3\\
          ~ & & LSTM     & 90.7 & 94.9 & 87.9 & 85.2 && 52.9 & 52.1 & 57.5 & 46.8 \\
		~ & & Bi-LSTM    & 86.5 & 94.5 & 90.6 & 91.4 && 56.2 & 46.3 & 53.5 & 50.4\\
        \cmidrule[0.5pt]{2-12}
        & \multirow{3}*{In-the-wild} 
        & Transformer    & 83.6 & 79.5 & 81.4 & 92.3 && 54.0 & 53.3 & 48.9 & 59.4\\
        ~ & & LSTM       & 84.5 & 80.8 & 84.6 & 82.7 && 50.5 & 55.0 & 50.2 & 50.2\\
		~ & & Bi-LSTM    & 87.2 & 85.2 & 87.5 & 87.2 && 54.4 & 51.7 & 50.2 & 49.8\\

        \midrule
        \multirow{6}*{0.6} 
        & \multirow{3}*{Clinical} 
          & Transformer  & 85.2 & 93.9 & 90.9 & 93.4 && 56.9 & 51.6 & 50.5 & 49.6\\
          ~ & & LSTM     & 86.3 & 94.4 & 87.9 & 85.9 && 57.3 & 51.8 & 50.0 & 50.0\\
		~ & & Bi-LSTM    & 84.6 & 93.9 & 90.6 & 93.9 && 54.3 & 54.0 & 50.4 & 50.9\\
        \cmidrule[0.5pt]{2-12}
        & \multirow{3}*{In-the-wild} 
        & Transformer    & 76.5 & 78.6 & 82.8 & 95.5 && 57.5 & 53.4 & 45.3 & 45.9\\
        ~ & & LSTM       & 82.9 & 76.6 & 83.4 & 84.2 && 49.8 & 49.4 & 45.9 & 45.9\\
		~ & & Bi-LSTM    & 76.8 & 85.2 & 85.4 & 90.4 && 57.7 & 45.9 & 45.9 & 45.9\\

        \midrule
        \multirow{6}*{0.7} 
        & \multirow{3}*{Clinical} 
          & Transformer  & 82.8 & 93.4 & 92.6 & 86.8 && 53.0 & 51.5 & 50.6 & 50.0\\
          ~ & & LSTM     & 87.0 & 94.9 & 81.0 & 85.2 && 53.8 & 50.6 & 50.0 & 50.0\\
		~ & & Bi-LSTM    & 79.5 & 94.0 & 90.6 & 93.3 && 53.8 & 52.2 & 50.0 & 50.0\\
        \cmidrule[0.5pt]{2-12}
        & \multirow{3}*{In-the-wild} 
        & Transformer    & 80.0 & 81.4 & 87.5 & 94.6 && 56.2 & 45.9 & 45.9 & 45.9\\
        ~ & & LSTM       & 83.5 & 77.2 & 83.4 & 84.2 && 49.2 & 48.5 & 45.9 & 45.9\\
		~ & & Bi-LSTM    & 76.8 & 87.1 & 86.6 & 86.9 && 48.3 & 45.9 & 45.9 & 45.9\\

        \midrule
        \multirow{6}*{0.8} 
        & \multirow{3}*{Clinical} 
          & Transformer  & 79.2 & 92.5 & 94.0 & 78.1 && 50.3 & 50.3 & 50.5 & 50.0\\
          ~ & & LSTM     & 85.8 & \textbf{95.8} & 82.3 & 85.2 && 51.6 & 50.2 & 50.0 & 50.0\\
		~ & & Bi-LSTM    & 78.6 & 93.4 & 90.6 & 85.8 && 52.3 & 49.8 & 50.0 & 50.0\\
        \cmidrule[0.5pt]{2-12}
        & \multirow{3}*{In-the-wild} 
        & Transformer    & 81.5 & 81.2 & 84.0 & 94.6 && 56.8 & 45.9 & 45.9 & 45.9\\
        ~ & & LSTM       & 86.3 & 78.4 & 86.0 & 78.7 && 52.0 & 45.9 & 45.9 & 45.9\\
		~ & & Bi-LSTM    & 78.1 & 87.1 & 86.5 & 86.3 && 33.8 & 45.9 & 45.9 & 45.9\\

        \midrule
        \multirow{6}*{0.9} 
        & \multirow{3}*{Clinical} 
          & Transformer  & 67.9 & 89.0 & 94.7 & 50.0 && 50.0 & 50.0 & 50.0 & 50.0\\
          ~ & & LSTM     & 57.8 & 92.2 & 75.1 & 50.0 && 50.0 & 50.0 & 50.0 & 50.0\\
		~ & & Bi-LSTM    & 75.5 & 90.1 & 90.9 & 67.1 && 50.0 & 50.0 & 50.0 & 50.0\\
        \cmidrule[0.5pt]{2-12}
        & \multirow{3}*{In-the-wild} 
        & Transformer    & 80.3 & 83.1 & 56.4 & 88.6 && 58.5 & 45.9 & 45.9 & 45.9\\
        ~ & & LSTM       & 72.9 & 82.0 & 89.8 & 82.5 && 45.9 & 45.9 & 45.9 & 45.9\\
		~ & & Bi-LSTM    & 80.0 & 70.8 & 77.2 & 73.0 && 45.9 & 45.9 & 45.9 & 45.9\\
        
        \bottomrule
    \end{tabular}}
    \label{tbl:resultMultiThre}
\end{table*}
}


%% file: paper.bbl
\begin{thebibliography}{36}
\expandafter\ifx\csname natexlab\endcsname\relax\def\natexlab#1{#1}\fi
\providecommand{\url}[1]{\texttt{#1}}
\providecommand{\href}[2]{#2}
\providecommand{\path}[1]{#1}
\providecommand{\DOIprefix}{doi:}
\providecommand{\ArXivprefix}{arXiv:}
\providecommand{\URLprefix}{URL: }
\providecommand{\Pubmedprefix}{pmid:}
\providecommand{\doi}[1]{\href{http://dx.doi.org/#1}{\path{#1}}}
\providecommand{\Pubmed}[1]{\href{pmid:#1}{\path{#1}}}
\providecommand{\bibinfo}[2]{#2}
\ifx\xfnm\relax \def\xfnm[#1]{\unskip,\space#1}\fi
\bibitem[{Benjasuwantep et~al.(2013)Benjasuwantep, Chaithirayanon and
  Eiamudomkan}]{benjasuwantep_feeding_2013}
\bibinfo{author}{Benjasuwantep, B.}, \bibinfo{author}{Chaithirayanon, S.},
  \bibinfo{author}{Eiamudomkan, M.}, \bibinfo{year}{2013}.
\newblock \bibinfo{title}{Feeding {Problems} in {Healthy} {Young} {Children}:
  {Prevalence}, {Related} {Factors} and {Feeding} {Practices}}.
\newblock \bibinfo{journal}{Pediatric Reports} \bibinfo{volume}{5},
  \bibinfo{pages}{e10}.
\newblock \URLprefix \url{https://www.mdpi.com/2036-7503/5/2/e10},
  \DOIprefix\doi{10.4081/pr.2013.e10}.
\bibitem[{Bolme et~al.(2010)Bolme, Beveridge, Draper and Lui}]{mossepaper}
\bibinfo{author}{Bolme, D.S.}, \bibinfo{author}{Beveridge, J.R.},
  \bibinfo{author}{Draper, B.A.}, \bibinfo{author}{Lui, Y.M.},
  \bibinfo{year}{2010}.
\newblock \bibinfo{title}{Visual object tracking using adaptive correlation
  filters}, in: \bibinfo{booktitle}{2010 IEEE Computer Society Conference on
  Computer Vision and Pattern Recognition}, pp. \bibinfo{pages}{2544--2550}.
\newblock \DOIprefix\doi{10.1109/CVPR.2010.5539960}.
\bibitem[{Carlin and Moon(2017)}]{carlin2017risk}
\bibinfo{author}{Carlin, R.F.}, \bibinfo{author}{Moon, R.Y.},
  \bibinfo{year}{2017}.
\newblock \bibinfo{title}{Risk factors, protective factors, and current
  recommendations to reduce sudden infant death syndrome: a review}.
\newblock \bibinfo{journal}{JAMA pediatrics} \bibinfo{volume}{171},
  \bibinfo{pages}{175--180}.
\bibitem[{Carreira and Zisserman(2017)}]{carreira2017quo}
\bibinfo{author}{Carreira, J.}, \bibinfo{author}{Zisserman, A.},
  \bibinfo{year}{2017}.
\newblock \bibinfo{title}{Quo vadis, action recognition? a new model and the
  kinetics dataset}, in: \bibinfo{booktitle}{proceedings of the IEEE Conference
  on Computer Vision and Pattern Recognition}, pp. \bibinfo{pages}{6299--6308}.
\bibitem[{Deng et~al.(2009)Deng, Dong, Socher, Li, Li and
  Fei-Fei}]{deng2009imagenet}
\bibinfo{author}{Deng, J.}, \bibinfo{author}{Dong, W.},
  \bibinfo{author}{Socher, R.}, \bibinfo{author}{Li, L.J.},
  \bibinfo{author}{Li, K.}, \bibinfo{author}{Fei-Fei, L.},
  \bibinfo{year}{2009}.
\newblock \bibinfo{title}{Imagenet: A large-scale hierarchical image database},
  in: \bibinfo{booktitle}{2009 IEEE conference on computer vision and pattern
  recognition}, \bibinfo{organization}{Ieee}. pp. \bibinfo{pages}{248--255}.
\bibitem[{Deng et~al.(2020)Deng, Guo, Ververas, Kotsia and
  Zafeiriou}]{retinaface2019}
\bibinfo{author}{Deng, J.}, \bibinfo{author}{Guo, J.},
  \bibinfo{author}{Ververas, E.}, \bibinfo{author}{Kotsia, I.},
  \bibinfo{author}{Zafeiriou, S.}, \bibinfo{year}{2020}.
\newblock \bibinfo{title}{Retinaface: Single-shot multi-level face localisation
  in the wild}, in: \bibinfo{booktitle}{Proceedings of the IEEE/CVF conference
  on computer vision and pattern recognition}, pp. \bibinfo{pages}{5203--5212}.
\bibitem[{Dutta and Zisserman(2019)}]{dutta2019vgg}
\bibinfo{author}{Dutta, A.}, \bibinfo{author}{Zisserman, A.},
  \bibinfo{year}{2019}.
\newblock \bibinfo{title}{The {VIA} annotation software for images, audio and
  video}, in: \bibinfo{booktitle}{Proceedings of the 27th ACM International
  Conference on Multimedia}, \bibinfo{publisher}{ACM}, \bibinfo{address}{New
  York, NY, USA}.
\newblock \URLprefix \url{https://doi.org/10.1145/3343031.3350535},
  \DOIprefix\doi{10.1145/3343031.3350535}.
\bibitem[{Farha and Gall(2019)}]{farha2019ms}
\bibinfo{author}{Farha, Y.A.}, \bibinfo{author}{Gall, J.},
  \bibinfo{year}{2019}.
\newblock \bibinfo{title}{Ms-tcn: Multi-stage temporal convolutional network
  for action segmentation}, in: \bibinfo{booktitle}{Proceedings of the IEEE/CVF
  Conference on Computer Vision and Pattern Recognition}, pp.
  \bibinfo{pages}{3575--3584}.
\bibitem[{Farneb{\"a}ck(2003)}]{farneback2003two}
\bibinfo{author}{Farneb{\"a}ck, G.}, \bibinfo{year}{2003}.
\newblock \bibinfo{title}{Two-frame motion estimation based on polynomial
  expansion}, in: \bibinfo{booktitle}{Scandinavian conference on Image
  analysis}, \bibinfo{organization}{Springer}. pp. \bibinfo{pages}{363--370}.
\bibitem[{Feichtenhofer(2020)}]{feichtenhofer2020x3d}
\bibinfo{author}{Feichtenhofer, C.}, \bibinfo{year}{2020}.
\newblock \bibinfo{title}{X3d: Expanding architectures for efficient video
  recognition}, in: \bibinfo{booktitle}{Proceedings of the IEEE/CVF conference
  on computer vision and pattern recognition}, pp. \bibinfo{pages}{203--213}.
\bibitem[{Gao et~al.(2021)Gao, Han, Li, Peng, Wang and
  Cheng}]{gao2021global2local}
\bibinfo{author}{Gao, S.H.}, \bibinfo{author}{Han, Q.}, \bibinfo{author}{Li,
  Z.Y.}, \bibinfo{author}{Peng, P.}, \bibinfo{author}{Wang, L.},
  \bibinfo{author}{Cheng, M.M.}, \bibinfo{year}{2021}.
\newblock \bibinfo{title}{Global2local: Efficient structure search for video
  action segmentation}, in: \bibinfo{booktitle}{Proceedings of the IEEE/CVF
  Conference on Computer Vision and Pattern Recognition}, pp.
  \bibinfo{pages}{16805--16814}.
\bibitem[{He et~al.(2016)He, Zhang, Ren and Sun}]{he2016deep}
\bibinfo{author}{He, K.}, \bibinfo{author}{Zhang, X.}, \bibinfo{author}{Ren,
  S.}, \bibinfo{author}{Sun, J.}, \bibinfo{year}{2016}.
\newblock \bibinfo{title}{Deep residual learning for image recognition}, in:
  \bibinfo{booktitle}{Proceedings of the IEEE conference on computer vision and
  pattern recognition}, pp. \bibinfo{pages}{770--778}.
\bibitem[{Huang et~al.(2019)Huang, Martens, Zimmerman and
  Ostadabbas}]{huang2019infant}
\bibinfo{author}{Huang, X.}, \bibinfo{author}{Martens, A.},
  \bibinfo{author}{Zimmerman, E.}, \bibinfo{author}{Ostadabbas, S.},
  \bibinfo{year}{2019}.
\newblock \bibinfo{title}{Infant contact-less non-nutritive sucking pattern
  quantification via facial gesture analysis.}, in: \bibinfo{booktitle}{CVPR
  Workshops}.
\bibitem[{Huber et~al.(2016)Huber, Hu, Tena, Mortazavian, Koppen, Christmas,
  Ratsch and Kittler}]{huber2016multiresolution}
\bibinfo{author}{Huber, P.}, \bibinfo{author}{Hu, G.}, \bibinfo{author}{Tena,
  R.}, \bibinfo{author}{Mortazavian, P.}, \bibinfo{author}{Koppen, P.},
  \bibinfo{author}{Christmas, W.J.}, \bibinfo{author}{Ratsch, M.},
  \bibinfo{author}{Kittler, J.}, \bibinfo{year}{2016}.
\newblock \bibinfo{title}{A multiresolution 3d morphable face model and fitting
  framework}, in: \bibinfo{booktitle}{Proceedings of the 11th International
  Joint Conference on Computer Vision, Imaging and Computer Graphics Theory and
  Applications}, \bibinfo{organization}{University of Surrey}.
\bibitem[{Idrees et~al.(2017)Idrees, Zamir, Jiang, Gorban, Laptev, Sukthankar
  and Shah}]{idrees_thumos_2017}
\bibinfo{author}{Idrees, H.}, \bibinfo{author}{Zamir, A.R.},
  \bibinfo{author}{Jiang, Y.G.}, \bibinfo{author}{Gorban, A.},
  \bibinfo{author}{Laptev, I.}, \bibinfo{author}{Sukthankar, R.},
  \bibinfo{author}{Shah, M.}, \bibinfo{year}{2017}.
\newblock \bibinfo{title}{The {THUMOS} {Challenge} on {Action} {Recognition}
  for {Videos} "in the {Wild}"}.
\newblock \bibinfo{journal}{Computer Vision and Image Understanding}
  \bibinfo{volume}{155}, \bibinfo{pages}{1--23}.
\newblock \URLprefix \url{http://arxiv.org/abs/1604.06182},
  \DOIprefix\doi{10.1016/j.cviu.2016.10.018}. \bibinfo{note}{arXiv:1604.06182
  [cs]}.
\bibitem[{Kataoka et~al.(2020)Kataoka, Wakamiya, Hara and
  Satoh}]{kataoka2020would}
\bibinfo{author}{Kataoka, H.}, \bibinfo{author}{Wakamiya, T.},
  \bibinfo{author}{Hara, K.}, \bibinfo{author}{Satoh, Y.},
  \bibinfo{year}{2020}.
\newblock \bibinfo{title}{Would mega-scale datasets further enhance
  spatiotemporal 3d cnns?}
\newblock \bibinfo{journal}{arXiv preprint arXiv:2004.04968} .
\bibitem[{Liu et~al.(2009)}]{liu2009beyond}
\bibinfo{author}{Liu, C.}, et~al., \bibinfo{year}{2009}.
\newblock \bibinfo{title}{Beyond pixels: exploring new representations and
  applications for motion analysis}.
\newblock Ph.D. thesis. Massachusetts Institute of Technology.
\bibitem[{Lucas et~al.(1981)Lucas, Kanade et~al.}]{lucas1981iterative}
\bibinfo{author}{Lucas, B.D.}, \bibinfo{author}{Kanade, T.}, et~al.,
  \bibinfo{year}{1981}.
\newblock \bibinfo{title}{An iterative image registration technique with an
  application to stereo vision}. volume~\bibinfo{volume}{81}.
\newblock \bibinfo{publisher}{Vancouver}.
\bibitem[{Maron and Lozano-P{\'e}rez(1997)}]{maron1997framework}
\bibinfo{author}{Maron, O.}, \bibinfo{author}{Lozano-P{\'e}rez, T.},
  \bibinfo{year}{1997}.
\newblock \bibinfo{title}{A framework for multiple-instance learning}.
\newblock \bibinfo{journal}{Advances in neural information processing systems}
  \bibinfo{volume}{10}.
\bibitem[{Martens et~al.(2020)Martens, Hines and
  Zimmerman}]{martens_changes_2020}
\bibinfo{author}{Martens, A.}, \bibinfo{author}{Hines, M.},
  \bibinfo{author}{Zimmerman, E.}, \bibinfo{year}{2020}.
\newblock \bibinfo{title}{Changes in non-nutritive suck between 3 and 12
  months}.
\newblock \bibinfo{journal}{Early Human Development} \bibinfo{volume}{149},
  \bibinfo{pages}{105141}.
\newblock \URLprefix
  \url{https://linkinghub.elsevier.com/retrieve/pii/S0378378220303807},
  \DOIprefix\doi{10.1016/j.earlhumdev.2020.105141}.
\bibitem[{McHugh(2012)}]{mchugh_interrater_2012}
\bibinfo{author}{McHugh, M.L.}, \bibinfo{year}{2012}.
\newblock \bibinfo{title}{Interrater reliability: the kappa statistic}.
\newblock \bibinfo{journal}{Biochemia Medica} \bibinfo{volume}{22},
  \bibinfo{pages}{276--282}.
\bibitem[{Medoff-Cooper and Ray(1995)}]{medoff-cooper_neonatal_1995}
\bibinfo{author}{Medoff-Cooper, B.}, \bibinfo{author}{Ray, W.},
  \bibinfo{year}{1995}.
\newblock \bibinfo{title}{Neonatal {Sucking} {Behaviors}}.
\newblock \bibinfo{journal}{Image: the Journal of Nursing Scholarship}
  \bibinfo{volume}{27}, \bibinfo{pages}{195--200}.
\newblock \URLprefix
  \url{https://onlinelibrary.wiley.com/doi/10.1111/j.1547-5069.1995.tb00858.x},
  \DOIprefix\doi{10.1111/j.1547-5069.1995.tb00858.x}.
\bibitem[{Oord et~al.(2016)Oord, Dieleman, Zen, Simonyan, Vinyals, Graves,
  Kalchbrenner, Senior and Kavukcuoglu}]{oord_wavenet_2016}
\bibinfo{author}{Oord, A.v.d.}, \bibinfo{author}{Dieleman, S.},
  \bibinfo{author}{Zen, H.}, \bibinfo{author}{Simonyan, K.},
  \bibinfo{author}{Vinyals, O.}, \bibinfo{author}{Graves, A.},
  \bibinfo{author}{Kalchbrenner, N.}, \bibinfo{author}{Senior, A.},
  \bibinfo{author}{Kavukcuoglu, K.}, \bibinfo{year}{2016}.
\newblock \bibinfo{title}{{WaveNet}: {A} {Generative} {Model} for {Raw}
  {Audio}}.
\newblock \URLprefix \url{http://arxiv.org/abs/1609.03499}.
  \bibinfo{note}{arXiv:1609.03499 [cs]}.
\bibitem[{Pock et~al.(2007)Pock, Urschler, Zach, Beichel and
  Bischof}]{pock2007duality}
\bibinfo{author}{Pock, T.}, \bibinfo{author}{Urschler, M.},
  \bibinfo{author}{Zach, C.}, \bibinfo{author}{Beichel, R.},
  \bibinfo{author}{Bischof, H.}, \bibinfo{year}{2007}.
\newblock \bibinfo{title}{A duality based algorithm for tv-l 1-optical-flow
  image registration}, in: \bibinfo{booktitle}{International Conference on
  Medical Image Computing and Computer-assisted Intervention},
  \bibinfo{organization}{Springer}. pp. \bibinfo{pages}{511--518}.
\bibitem[{Poore et~al.(2008)Poore, Zimmerman, Barlow, Wang and
  Gu}]{poore2008patterned}
\bibinfo{author}{Poore, M.}, \bibinfo{author}{Zimmerman, E.},
  \bibinfo{author}{Barlow, S.}, \bibinfo{author}{Wang, J.},
  \bibinfo{author}{Gu, F.}, \bibinfo{year}{2008}.
\newblock \bibinfo{title}{Patterned orocutaneous therapy improves sucking and
  oral feeding in preterm infants}.
\newblock \bibinfo{journal}{Acta paediatrica} \bibinfo{volume}{97},
  \bibinfo{pages}{920--927}.
\bibitem[{Psaila et~al.(2017)Psaila, Foster, Pulbrook and
  Jeffery}]{psaila_infant_2017}
\bibinfo{author}{Psaila, K.}, \bibinfo{author}{Foster, J.P.},
  \bibinfo{author}{Pulbrook, N.}, \bibinfo{author}{Jeffery, H.E.},
  \bibinfo{year}{2017}.
\newblock \bibinfo{title}{Infant pacifiers for reduction in risk of sudden
  infant death syndrome}.
\newblock \bibinfo{journal}{Cochrane Database of Systematic Reviews}
  \bibinfo{volume}{2017}.
\newblock \URLprefix \url{http://doi.wiley.com/10.1002/14651858.CD011147.pub2},
  \DOIprefix\doi{10.1002/14651858.CD011147.pub2}.
\bibitem[{Shi et~al.(1994)}]{shi1994good}
\bibinfo{author}{Shi, J.}, et~al., \bibinfo{year}{1994}.
\newblock \bibinfo{title}{Good features to track}, in: \bibinfo{booktitle}{1994
  Proceedings of IEEE conference on computer vision and pattern recognition},
  \bibinfo{organization}{IEEE}. pp. \bibinfo{pages}{593--600}.
\bibitem[{Teed and Deng(2020)}]{teed2020raft}
\bibinfo{author}{Teed, Z.}, \bibinfo{author}{Deng, J.}, \bibinfo{year}{2020}.
\newblock \bibinfo{title}{Raft: Recurrent all-pairs field transforms for
  optical flow}, in: \bibinfo{booktitle}{European conference on computer
  vision}, \bibinfo{organization}{Springer}. pp. \bibinfo{pages}{402--419}.
\bibitem[{Vaswani et~al.(2017)Vaswani, Shazeer, Parmar, Uszkoreit, Jones,
  Gomez, Kaiser and Polosukhin}]{vaswani2017attention}
\bibinfo{author}{Vaswani, A.}, \bibinfo{author}{Shazeer, N.},
  \bibinfo{author}{Parmar, N.}, \bibinfo{author}{Uszkoreit, J.},
  \bibinfo{author}{Jones, L.}, \bibinfo{author}{Gomez, A.N.},
  \bibinfo{author}{Kaiser, {\L}.}, \bibinfo{author}{Polosukhin, I.},
  \bibinfo{year}{2017}.
\newblock \bibinfo{title}{Attention is all you need}.
\newblock \bibinfo{journal}{Advances in neural information processing systems}
  \bibinfo{volume}{30}.
\bibitem[{Wan et~al.(2022)Wan, Zhu, Luan, Gulati, Huang, Schwartz-Mette, Hayes,
  Zimmerman and Ostadabbas}]{wan_infanface_2022}
\bibinfo{author}{Wan, M.}, \bibinfo{author}{Zhu, S.}, \bibinfo{author}{Luan,
  L.}, \bibinfo{author}{Gulati, P.}, \bibinfo{author}{Huang, X.},
  \bibinfo{author}{Schwartz-Mette, R.}, \bibinfo{author}{Hayes, M.},
  \bibinfo{author}{Zimmerman, E.}, \bibinfo{author}{Ostadabbas, S.},
  \bibinfo{year}{2022}.
\newblock \bibinfo{title}{{InfAnFace}: {Bridging} the {Infant}--{Adult}
  {Domain} {Gap} in {Facial} {Landmark} {Estimation} in the {Wild}}.
\newblock \bibinfo{journal}{26th International Conference on Pattern
  Recognition (ICPR)} .
\bibitem[{Yi et~al.(2021)Yi, Wen and Jiang}]{yi2021asformer}
\bibinfo{author}{Yi, F.}, \bibinfo{author}{Wen, H.}, \bibinfo{author}{Jiang,
  T.}, \bibinfo{year}{2021}.
\newblock \bibinfo{title}{Asformer: Transformer for action segmentation}.
\newblock \bibinfo{journal}{arXiv preprint arXiv:2110.08568} .
\bibitem[{Yue-Hei~Ng et~al.(2015)Yue-Hei~Ng, Hausknecht, Vijayanarasimhan,
  Vinyals, Monga and Toderici}]{yue2015beyond}
\bibinfo{author}{Yue-Hei~Ng, J.}, \bibinfo{author}{Hausknecht, M.},
  \bibinfo{author}{Vijayanarasimhan, S.}, \bibinfo{author}{Vinyals, O.},
  \bibinfo{author}{Monga, R.}, \bibinfo{author}{Toderici, G.},
  \bibinfo{year}{2015}.
\newblock \bibinfo{title}{Beyond short snippets: Deep networks for video
  classification}, in: \bibinfo{booktitle}{Proceedings of the IEEE conference
  on computer vision and pattern recognition}, pp. \bibinfo{pages}{4694--4702}.
\bibitem[{Zavala~Abed et~al.(2020)Zavala~Abed, Oneto, Abreu and
  Chediak}]{zavala_abed_how_2020}
\bibinfo{author}{Zavala~Abed, B.}, \bibinfo{author}{Oneto, S.},
  \bibinfo{author}{Abreu, A.R.}, \bibinfo{author}{Chediak, A.D.},
  \bibinfo{year}{2020}.
\newblock \bibinfo{title}{How might non nutritional sucking protect from sudden
  infant death syndrome}.
\newblock \bibinfo{journal}{Medical Hypotheses} \bibinfo{volume}{143},
  \bibinfo{pages}{109868}.
\newblock \URLprefix
  \url{https://linkinghub.elsevier.com/retrieve/pii/S0306987720307386},
  \DOIprefix\doi{10.1016/j.mehy.2020.109868}.
\bibitem[{Zhu et~al.(2023)Zhu, Wan, Hatamimajoumerd, Kamath, Jain, Zlota,
  Grace, Rowan, Goodwin, Schwartz-Mette, Zimmerman, Hayes and
  Ostadabbas}]{zhu2023video}
\bibinfo{author}{Zhu, S.}, \bibinfo{author}{Wan, M.},
  \bibinfo{author}{Hatamimajoumerd, E.}, \bibinfo{author}{Kamath, C.V.},
  \bibinfo{author}{Jain, K.}, \bibinfo{author}{Zlota, S.},
  \bibinfo{author}{Grace, E.}, \bibinfo{author}{Rowan, C.},
  \bibinfo{author}{Goodwin, M.}, \bibinfo{author}{Schwartz-Mette, R.},
  \bibinfo{author}{Zimmerman, E.}, \bibinfo{author}{Hayes, M.},
  \bibinfo{author}{Ostadabbas, S.}, \bibinfo{year}{2023}.
\newblock \bibinfo{title}{A video-based end-to-end pipeline for non-nutritive
  sucking action recognition and segmentation in young infants}, in:
  \bibinfo{booktitle}{International Conference on Medical Image Computing and
  Computer-Assisted Intervention (MICCAI)}.
\bibitem[{Zimmerman et~al.(2020)Zimmerman, Carpenito and
  Martens}]{zimmerman_changes_2020}
\bibinfo{author}{Zimmerman, E.}, \bibinfo{author}{Carpenito, T.},
  \bibinfo{author}{Martens, A.}, \bibinfo{year}{2020}.
\newblock \bibinfo{title}{Changes in infant non-nutritive sucking throughout a
  suck sample at 3-months of age}.
\newblock \bibinfo{journal}{PLOS ONE} \bibinfo{volume}{15},
  \bibinfo{pages}{e0235741}.
\newblock \URLprefix \url{https://dx.plos.org/10.1371/journal.pone.0235741},
  \DOIprefix\doi{10.1371/journal.pone.0235741}.
\bibitem[{Zimmerman and Foran(2017)}]{zimmerman_patterned_2017}
\bibinfo{author}{Zimmerman, E.}, \bibinfo{author}{Foran, M.},
  \bibinfo{year}{2017}.
\newblock \bibinfo{title}{Patterned auditory stimulation and suck dynamics in
  full-term infants}.
\newblock \bibinfo{journal}{Acta Paediatrica} \bibinfo{volume}{106},
  \bibinfo{pages}{727--732}.
\newblock \URLprefix
  \url{https://onlinelibrary.wiley.com/doi/10.1111/apa.13751},
  \DOIprefix\doi{10.1111/apa.13751}.

\end{thebibliography}
